\documentclass[sigconf, nonacm]{acmart}

\usepackage[version=4]{mhchem}
\usepackage{chemfig}

\usepackage{forest}
\graphicspath{{./figures/}}

\usepackage{bbm}

\author{Dominik Köhler}
\email{dominik.koehler@upb.de}
\affiliation{%
  \institution{Paderborn University}
  \city{Paderborn}
  \country{Germany}
}

\author{Stefan Heindorf}
\email{heindorf@upb.de}
\affiliation{%
  \institution{Paderborn University}
  \city{Paderborn}
  \country{Germany}
}

\usepackage{algorithm}
\usepackage{algpseudocode}

\newlength\myindent
\algnewcommand{\algvar}{\texttt}
\algnewcommand{\assign}{\leftarrow}
\algnewcommand{\NULL}{\textsc{null}}

\newcommand{\aggr}{\text{aggr}}

\newcommand{\cEL}{\ensuremath{\mathcal{EL}}}

\definecolor{ownred}{HTML}{E15759}
\definecolor{owngreen}{HTML}{59A14F}
\definecolor{ownyellow}{HTML}{EDC948}
\definecolor{owngrey}{HTML}{BAB0AC}
\definecolor{ownorange}{HTML}{F28E2B}
\definecolor{ownblue}{HTML}{4E79A7}

\definecolor{tab20darkblue}{HTML}{4e79a7}
\definecolor{tab20darkgreen}{HTML}{59a14f}
\definecolor{tab20darkred}{HTML}{e15759}
\definecolor{tab20darkorange}{HTML}{f28e2b}
\definecolor{tab20darkturquoise}{HTML}{499894}
\definecolor{tab20darkgray}{HTML}{79706e}
\definecolor{tab20darkbrown}{HTML}{9d7660}
\definecolor{tab20darkpurple}{HTML}{b07aa1}
\definecolor{tab20darkpink}{HTML}{e377c2}
\definecolor{tab20lightgreen}{HTML}{8cd17d}
\definecolor{tab20lightblue}{HTML}{a0cbd8}
\definecolor{tab20lightorange}{HTML}{ffbb78}
\definecolor{tab20lightred}{HTML}{ff9896}
\definecolor{tab20lightpurple}{HTML}{c5b0d5}
\definecolor{tab20lightbrown}{HTML}{c49c94}
\definecolor{tab20lightpink}{HTML}{f7b6d2}
\definecolor{tab20lightgray}{HTML}{c7c7c7}
\definecolor{tab20lightturquoise}{HTML}{9edae5}
\definecolor{tab20lightyellow}{HTML}{dbdb8d}
\definecolor{tab20darkyellow}{HTML}{bcbd22}
\definecolor{HouseType0}{RGB}{89, 161, 79}
\definecolor{HouseType1}{RGB}{242, 142, 43}
\definecolor{HouseType2}{RGB}{78, 121, 167}
\definecolor{HouseType3}{RGB}{225, 87, 89}

\begin{document}

\algrenewcommand\algorithmicindent{1.1em}

\title{Utilizing Description Logics for Global Explanations of Heterogeneous Graph Neural Networks}

\begin{abstract}
Graph Neural Networks (GNNs) are effective for node classification in graph-structured data, but they lack explainability, especially at the global level. Current research mainly utilizes subgraphs of the input as local explanations or generates new graphs as global explanations. However, these graph-based methods are limited in their ability to explain classes with multiple sufficient explanations. To provide more expressive explanations, we propose utilizing class expressions (CEs) from the field of description logic (DL). Our approach explains heterogeneous graphs with different types of nodes using CEs in the \cEL{} description logic. To identify the best explanation among multiple candidate explanations, we employ and compare two different scoring functions: (1) For a given CE, we construct multiple graphs, have the GNN make a prediction for each graph, and aggregate the predicted scores. (2) We score the CE in terms of fidelity, i.e., we compare the predictions of the GNN to the predictions by the CE on a separate validation set. Instead of subgraph-based explanations, we offer CE-based explanations.
\end{abstract}

\maketitle

\section{Introduction}

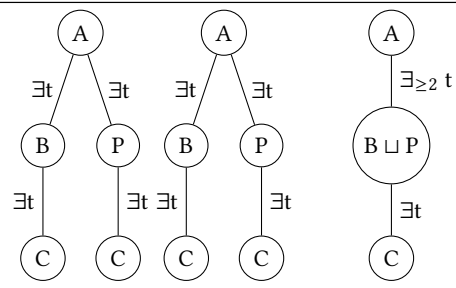
\begin{figure}[tb]
    \centering
    \begin{tabular}{@{}l@{\hskip -77pt}l@{\hskip 20pt}c@{}}
\toprule
\textbf{Fraudulent assets in a financial network} &  \\
\midrule
\hspace{30pt}
\begin{tikzpicture}
  \node[circle, draw] (A) at (0,2) {A};
  \node[circle, draw] (B) at (-0.5,0.5) {B};
  \node[circle, draw] (C) at (0.5,0.5) {P};
  \node[circle, draw] (P) at (-0.5,-1) {C};
  \node[circle, draw] (Pt) at (0.5,-1) {C};
  
  \draw (A) --node[left] {$\exists$t} (B);
  \draw (A) --node[right] {$\exists$t} (C);
  \draw (C) --node[right] {$\exists$t} (Pt);
  \draw (B) --node[left] {$\exists$t} (P);
\end{tikzpicture}
&
\begin{tikzpicture}
  \node[circle, draw] (A) at (-2,2) {A};
  \node[circle, draw] (B) at (-2.5,0.5) {B};
  \node[circle, draw] (C) at (-1.5,0.5) {P};
  \node[circle, draw] (P) at (-2.5,-1) {C};
  \node[circle, draw] (Pt) at (-1.5,-1) {C};
  
  \draw (A) --node[left] {$\exists$t} (B);
  \draw (A) --node[right] {$\exists$t} (C);
  \draw (C) --node[right] {$\exists$t} (Pt);
  \draw (B) --node[left] {$\exists$t} (P);
\end{tikzpicture}
&
\begin{tikzpicture}
  \node[circle, draw] (A) at (-5,2) {A};
  \node[circle, draw] (B) at (-5,0.5) {B $\sqcup$ P};
  \node[circle, draw] (C) at (-5,-1) {C};
  
  \draw (A) --node[right] {$\exists_{\ge2}$ t} (B);
  \draw (B) --node[right] {$\exists$t} (C);
\end{tikzpicture}
\hspace{30pt}
\\
\bottomrule
\end{tabular}
        \caption{In a financial network, where each node type is a specific account, one asset account A might be labeled as fraudulent if it has transactions t with at least two accounts of type business or personal (nodes of type B or P), which both transact money to (possibly distinct) known criminal accounts (nodes of type C).
        The two left-hand graphs refers to the CE $\text{A} \sqcap \exists \text{t.} \left(\text{B} \sqcap  \exists \text{t.C} \right) \sqcap \exists\text{t.}\left( \text{P} \sqcap \exists \text{t.C} \right)$ in $\cEL$, which we implemented in our approach, whereas the class expression (CE) on the right side refers to the more accurate CE $\text{A} \exists_{\ge 2} \text{t. ((B $\sqcup$ P) $\exists$C) }$, an approach we left for future work.
        }
    \label{fig:Finance}
\end{figure}

Node classification in graphs such as knowledge graphs (KGs), product graphs, protein graphs, or citation graphs, is a critical task often addressed using graph neural networks (GNNs)~\cite{Hu2020Open}. However, understanding their predictions, which is crucial for many applications, is challenging. Hence, several explanation approaches such as GNNExplainer~\cite{Ying2019GNNExplainer}, PGExplainer~\cite{Luo2020PGExplainer}, SubgraphX~\cite{Yuan2021SubgraphX}, XGNN~\cite{Yuan2020XGNN}, and GNNInterpreter~\cite{Wang2023GNNInterpreter} have been proposed, which can be categorized into two types: \emph{local} and \emph{global} explanations. \emph{Local} explanations focus on why a particular node is classified in a certain way, often represented by subgraphs highlighting crucial parts for that node's prediction. \emph{Global} explanations, on the other hand, elucidate why specific labels are predicted. They do so in terms of graph patterns that describe the set of nodes having a certain label. Typically, a graph pattern is a subgraph of the metagraph, which encompasses node types and their connections.

Contemporary global explainers, however, do not offer sufficient expressiveness. Whereas common families of GNNs can learn intricate patterns expressed in the description logic $\mathcal{ALCQ}$~\cite{Barcelo2020Logical} supporting conjunctions, disjunctions, negations, and cardinality restrictions, a subgraph of a metagraph is unable to capture disjunctions, negations, and cardinality restrictions. Figure~\ref{fig:Finance} shows an example. In this example, we assume that the asset account $A$ is fraudulent if and only if it transacts money to at least two different accounts (which can be business or private accounts) that in turn send money to a criminal account $C$ (either the same criminal account or a different one). The ideal class expression to describe this situation is shown on the right. However, a single subgraph explanation cannot describe this situation, as it can neither express cardinality restrictions ($\exists_{\geq 2}$) nor disjunctions ($B \sqcup P$).

In response, we propose explaining GNNs based on $\cEL$~\cite{Baader2003DLHandbook} class expressions (CEs) and in future work on $\mathcal{ALCQ}$, with each node type as a named class and each edge type as an object property. Then a CE can directly be used as a binary classifier. For example, given the CE 
$A \exists_{\ge 2} \text{t. ((B $\sqcup$ P)$\exists$ C }$, we can check whether the node fulfills this CE for each node in the graph. 

Our approach works as follows: We randomly generate CEs via beam search and select the CE maximizing our scoring function, making the explanation independent of the dataset~\cite{Ramaswamy2023Overlooked}. We propose two scoring functions: (1) We use the CE as a binary classifier and score its fidelity on a validation set. (2) We choose the CE that maximizes the prediction of the GNN. More precisely, given a CE $ce$ that aims to explain the prediction of label $l$, we synthesize random graphs $G(ce)$ from the CE. At the same time, for each graph $g \in G(ce)$, we synthesize a node $v_g$. Then we evaluate the GNN for this graph, node, and label, denoted as $f_l(g,v_g)$. To retrieve the best explanatory CE $ce'$, we average the GNN output for all synthesized graphs, resulting in the formula: 
\begin{align*}\label{eq:score}
    ce' = \arg\max_{ce}[\aggr_{g \in G(ce)} f_l(g, v_g)].
\end{align*}
Typical aggregation functions include \emph{mean} and \emph{max}.

To evaluate our approach, we construct a heterogeneous version of the \textsc{BA-Shapes} dataset~\cite{Ying2019GNNExplainer} and as a first step, implement an explainer for the description logic $\mathcal{EL}$~\cite{Baader2003DLHandbook}, which is a subset of $\mathcal{ALCQ}$. We adapt metrics from graph-based explanations to evaluate CE-based explanations. Our main contribution is a novel approach that utilizes CEs to explain GNNs globally. Our code is publicly available.%
\footnote{\url{https://github.com/ds-jrg/xgnn-dl}}

\section{Related Work}

We briefly review related works on GNNs, explanation approaches for GNNs, and learning class expressions. 

\subsection{Graph Neural Networks}
GNNs are employed for tasks such as node classification, graph classification, and link prediction~\cite{Thomas2023GAINGNNSurvey,Wu2021ComprSurvey,Ye2022Survey}. They commonly utilize the message passing framework~\cite{Wang2022How,Gilmer2017Neural}: In each layer, the GNN aggregates node features from adjacent nodes for each node $v$. In a GNN with $n$ layers, the $n$-hop environment of $v$ influences its embedding. Typically GNNs operate in the \emph{transductive} setting, i.e., the same nodes are used for training and testing, e.g., RGCN~\cite{Schlichtkrull2017RGCN}. However, there is growing support for the \emph{inductive} setting where training and testing nodes can differ, e.g., GraphSAGE~\cite{Hamilton2017GraphSAGE}. In this paper, we focus on inductive GNNs.

\subsection{Explaining GNNs}
Explaining GNNs has been widely studied in the literature, as summarized in the surveys by~\citet{Yuan2023Explainability} and \citet{LI2022ExGNN_Survey}. Evaluation frameworks for the explanations have been proposed by \citet{Agarwal2022Evaluating} and \citet{Longa2022Explaining}. Approaches can be divided as to whether they produce \emph{local} or \emph{global} explanations.

\emph{Local explanations}, also known as instance-level explanations, explain the prediction of a \emph{single node}. The typical outputs are small, important subgraphs of the original graph containing the node to be explained. The first approach was GNNExplainer~\cite{Ying2019GNNExplainer}, which maximizes the mutual information between a GNN's prediction and the explanatory subgraph. PG-Explainer~\cite{Luo2020PGExplainer} and GEM~\cite{Lin2021GEM} synthesize explanatory subgraphs via neural networks. SubgraphX~\cite{Yuan2021SubgraphX} searches for an explanatory subgraph via Monte-Carlo Tree Search (MCTS). GStarX~\cite{Zhang2022GStarX} employs a modified version of Shapley values, taking the graph structure into account for selecting the collaborations. All of the aforementioned \emph{local} explainers except for PGExplainer operate in the transductive setting, i.e., predictions and explanations can only be computed for a graph seen during training, but not for new, unseen graphs. Moreover, they can only explain a single node prediction in contrast to global explainers.

\emph{Global explanations}, also known as model-level explanations, explain why \emph{multiple} nodes are predicted to have a \emph{certain label}. An explanation is typically a subgraph of the metagraph. XGNN~\cite{Yuan2020XGNN} builds an explanatory graph via reinforcement learning, optimizing the GNN score for the created graph. Similarly, GNNInterpreter~\cite{Wang2023GNNInterpreter} searches for graphs with a high GNN score, which are similar to the original dataset. PAGE~\cite{Shin2022PAGEProto} clusters graphs to obtain a central graph that explains all graphs in the cluster. Most closely related to our work, GLGExplainer~\cite{Azzolin2022Global} generates explanations as Boolean combinations of graphical graph patterns. The graphical graph patterns represent prototypical explanations obtained from a \emph{local} explainer. To create \emph{global} predictions with the \emph{local} explanations, the paper considers the similarity between local graph patterns and graphs in an embedding space (via so-called concept projections). As the embedding space is not amenable to humans, humans can only use graphical explanations to produce approximate and intuitive predictions on new data. In contrast, our CEs have clear and formal semantics that humans can easily use to make precise, unambiguous predictions on new data. 

One other method that combines sub-symbolic and symbolic techniques to generate global explanations is described by~\citet{Himmelhuber2021Combining,Himmelhuber2021Concept}. This method employs a local subgraph-based explainer, e.g. GNNExplainer~\cite{Ying2019GNNExplainer} to construct an ontology, which in turn is used by a class expression learner to create CE-based global explanations. This method most closely resembles our approach. However, we directly create global explanations bypassing the need to create local explanations. Moreover, we generate new graphs rather than relying on existing local explanations, which allows us to recognize structures learned by the GNN that are not present in the original dataset.

\subsection{Learning Class Expressions}

Description Logics~\cite{Baader2003DLHandbook} are a family of formal languages for representing and reasoning about structured knowledge with varying degrees of expressiveness. For example, the web ontology language (OWL) corresponds to $\mathcal{SROIQ(D)}$, whereas our approach focuses on the lightweight DL $\cEL$ (see Section~\ref{sec:preliminaries}). Several approaches for learning CEs from positive and negative examples in knowledge bases have been proposed. For example, Evolearner~\cite{Heindorf2022EvoLearner} searches for CEs via evolutionary algorithms, DRILL~\cite{Demir2023Neuro-Symbolic} searches for them via reinforcement learning, CLIP~\cite{Kouagou2022Learning} speeds up the search by predicting the length of a CE, OntoSample~\cite{Baci2023Accelerating} by sampling the ontology, and NCES~\cite{Kouagou2023NCES, Kouagou2023NCES2} as well as ROCES~\cite{Kouagou2024ROCES} directly synthesize CEs via set transformers. However, none of the approaches have been used for explaining GNNs.

\section{Approach}

After a brief introduction of preliminaries, we describe our search algorithm.

\subsection{Preliminaries}\label{sec:preliminaries}
Our goal is to explain GNNs trained on heterogeneous graphs that allow multiple node and edge types~\citep{Shi2022Heterogeneous, wang2023enabling}.
\begin{definition}\label{def:hgraph}
A \textit{heterogeneous graph} $G$ is formally defined as a tuple $G := (V, E, T_V, T_E, N_V, N_E, \rho, \phi_V, \phi_E)$, where
\begin{itemize}
        \item \( V \) is the set of nodes,
        \item \( E \) is the set of edges with $V \cap E \neq \emptyset$,
        \item \( T_V \) is the set of node types,
        \item \( T_E \) is the set of edge types,
        \item $\rho:E \rightarrow V \times V$ is a total function mapping each edge to its pair of nodes,
        \item \( \phi_V: V \rightarrow T_V \) is a total function mapping each node to its type,
        \item \( \phi_E: E \rightarrow T_E \) is a total function mapping each edge to its type.
        \item We assume that each node type and edge type has a unique name, denoted by $N_V$ and $N_E$, respectively.
\end{itemize}
\end{definition}

The definition corresponds to a typed multi-graph, where each node and each edge has a single type and there can be multiple edges between the same pair of nodes.

\subsubsection{Description logic}

In description logics, \emph{individuals} are instances of \textit{named classes} and \emph{links} between individuals are instances of \textit{properties}. In the terminology of heterogeneous graphs, \emph{nodes} are instances of \emph{node types} and \emph{edges} are instances of \emph{edge types}.

\begin{figure}
\centering
\begin{table}[H]
  \centering
  \setlength{\tabcolsep}{3pt}
  \caption{Syntax and semantics of $\mathcal{EL}$. $\mathcal{I}=(\Delta^I, \cdot^\mathcal{I})$ is an interpretation where $\Delta^\mathcal{I}$ is its domain and $\cdot^\mathcal{I}$ is the interpretation function.}
  \label{tab:description-logic}
  
  \resizebox{\columnwidth}{!}{
  \begin{tabular}{@{}lll@{}}
  \toprule
  \textbf{Syntax} & \textbf{Construct} & \textbf{Semantics}\\
  \midrule
  $r$           & property & $r^\mathcal{I}\subseteq \Delta^\mathcal{I}\times \Delta^\mathcal{I}$\\
  $C, D$ & classes & $C^{\mathcal{I}}\subseteq{\Delta^\mathcal{I}}$, $D^{\mathcal{I}}\subseteq{\Delta^\mathcal{I}}$\\
  \midrule
  $C \sqcap D$  & intersection & $C^\mathcal{I}\cap D^\mathcal{I}$\\
  $\exists\ r.C$ & existential restriction & $\{a^\mathcal{I} \in \Delta^\mathcal{I}|\ \exists~ b^\mathcal{I} \in C^\mathcal{I}, (a^\mathcal{I},b^\mathcal{I})\in r^\mathcal{I}\}$\\
  \bottomrule
  \end{tabular}}
\end{table}
\end{figure}

We define the syntax and semantics of class expressions over heterogeneous graphs as shown in Table~\ref{tab:description-logic}.

\paragraph{Syntax} Each named class $A, B \in N_V$ corresponds to a node type and each property $r \in N_E$ corresponds to an edge type. A complex class expression $C$ can be built from basic constructs: If $A$ is a named class, then $A$ is a class expression; If $C, D$ are class expressions and $r$ is a property, then both $C\sqcap D$ as well as $\exists r.C$ are class expressions. 

\paragraph{Semantics} Given a (complex) class expression $C$, we want to define all nodes $C^\mathcal{I}$ that fulfill this class expression. If $C$ is a named class, its instances $C^\mathcal{I}$ is the set of nodes with the node type name $C$. If $C$ is a complex class expression, its instances $C^\mathcal{I}$ can be obtained by recursively applying the semantics of its constructs. This is similar to a retrieval operator~\cite{Lehmann2010PhD}, but we will call the mapping a \textit{homomorphism}. A node $v$ in a graph \textit{fulfills} a CE if a homomorphism from the CE to the graph exists, where the CE's root class is mapped onto the node $v$.

In an intersection $A \sqcap B$, $A$ and $B$ are called operands; in an existential restriction $\exists r. B$, $B$ is referred to as filler and $r$ to the property. When it is clear from the context, we may use the terms named class, class, node type, and node type name; and also property, edge type, and edge type name interchangeably. Our definition allows intersections and existential restrictions, resembling the description logic $\cEL$~\citep{Baader2003DLHandbook}.

\subsection{The Search Algorithm}

\begin{figure}
    \centering
    
\begin{tikzpicture}[node distance=2cm, auto]
  \node[draw] (CE) {CEs};
  \node[draw] (Graph) [right of=CE, node distance = 4cm] {create graphs from CEs};
  \node[draw] (Score) [below of=Graph, node distance = 1.5cm] {score graphs};
  \node[draw] (Best) [below of=Score, node distance = 1.5cm] {select best CEs};
  \node[draw] (Mutate) [left of=Best, node distance = 4cm] {mutate CE};

  \draw[->] (CE) -- (Graph);
  \draw[->] (Graph) -- (Score);
  \draw[->] (Score) -- (Best);
  \draw[->] (Mutate) -- (CE);
  \draw[->] (Best) -- (Mutate);
  \draw[->] (Best) -- (CE);
\end{tikzpicture}
\caption{An overview of our approach: We start with class expressions (CEs),  and score them individually by using the GNN on graphs that have a node fulfilling the CE. For the next iteration, we take (1) the best results and (2) mutated versions of the best results.}\label{fig:overview}
\end{figure}
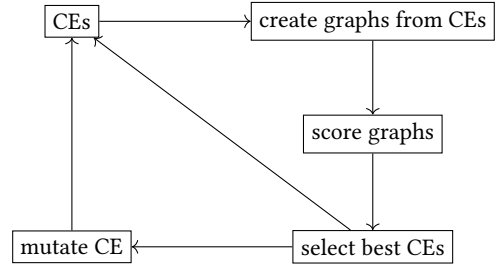

Figure~\ref{fig:overview} gives an overview of our approach illustrating our search for the best explanatory class expression, for the predictions of a given GNN. We do so via beam search: Initially, we randomly generate class expressions. To evaluate the suitability of a class expression, we randomly generate graphs with a node fulfilling the class expression and score the graphs and thus the CE via the trained GNN. In the next iteration, we continue with the best CEs found and their mutations.

\paragraph{Beam search}
Algorithm~\ref{alg:beam-search} shows the pseudocode of our beam search. As parameters, it takes the number of iterations $n$, the beam width $k$, the set of all available $classes$, and the $class$ to be explained. We generate CEs of (increasing) length for $n$ iterations: Initially, we create $k$ CEs (line 4). In each iteration, for each candidate, we add a mutation of this candidate (lines 7--10). This is done by adding another CE of the form $\exists r. CLASS$. From the resulting $2k$ candidates, we again choose the best $k$ candidates by a scoring function for the next iteration (lines 12-13).

In the following, we presume that all edges have the same type, but the approach is extendable to datasets with various edge types.

\begin{algorithm}[H]
\caption{Beam Search}\label{alg:beam-search}
\begin{algorithmic}[1]
\Procedure{BeamSearch}{$n$, $k$, $classes$, $class$}
\State $candidates \gets $ EmptyList()
\For{$1 \le i \le k$}
\State $candidates$.append(\textsc{RandomCE}($classes, class$))
\EndFor
\For{$1 \le i \le n$}
    \For{$ce$ in $candidates$}
        \State $ce_{new} \gets \textsc{MutateCE}(ce, classes)$
        \State $ce_{new}$.score $\gets$ \textsc{ScoreCE}($ce_{new}, class$)
        \State $candidates$.append($ce_{new}$)
    \EndFor
    \State $\triangleright$ Choose the best $k$ CEs from $candidates$:
    \State $candidates \gets $\textsc{Sort}$(candidates, score)[:k]$
\EndFor
\State \textbf{return} $candidates$
\EndProcedure
\end{algorithmic}
\end{algorithm}

\paragraph{Create random CE}
For creating a random CE (line 4 of Algorithm~\ref{alg:beam-search}, we are given the $class$ to explain and all available $classes$. We uniformly randomly pick a class $class_{new}$ from $classes$ and create the CE $class \sqcap \exists r.class_{new}$. For example, with $class=A$, and $classes=\{A, B, C\}$, we might randomly create the CE $A \sqcap \exists r.B$.

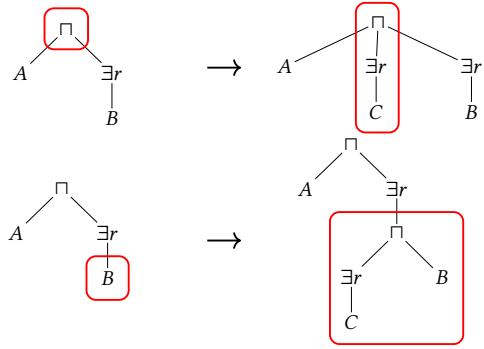
\begin{figure}[ht]
\centering
\scalebox{0.9}{
\begin{minipage}{0.20\textwidth}
\centering
\begin{forest}
for tree={s sep=10mm, inner sep=1, l=1}
  [
    {$\sqcap$},
    name=root,
    tikz={\node [draw=red, thick, rounded corners, fit=(), inner xsep=5pt, inner ysep=5pt] {};}
    [$A$]
    [$\exists r$
      [$B$]
    ]
  ]
\end{forest}
\end{minipage}
\begin{minipage}{0.05\textwidth}
\centering
\tikz{\draw[->, thick] (0,0)--+(0.5,0);}
\end{minipage}
\begin{minipage}{0.20\textwidth}
\centering
\begin{forest}
for tree={s sep=10mm, inner sep=1, l=1}
  [
    {$\sqcap$},
    name=root,
    tikz={\node [draw=red, thick, rounded corners, fit=() (newNode), inner xsep=4.5pt, inner ysep=4.5pt] {};}
    [$A$]
    [$\exists r$ [$C$, name=newNode]]
    [$\exists r$
      [$B$]
    ]
  ]
\end{forest}
\end{minipage}}
\scalebox{0.9}{
\begin{minipage}{0.20\textwidth}
\centering
\begin{forest}
for tree={s sep=10mm, inner sep=1, l=1}
 [
  $\sqcap$
    [$A$]
    [$\exists r$, name=start
      [$B$, name=end, tikz={%
   \node [draw=red, thick, rounded corners, fit=(), inner xsep=5pt, inner ysep=5pt] {};
   }
        ]
      ]
    ]
  ]
\end{forest}
\end{minipage}
\begin{minipage}{0.05\textwidth}
\centering
\tikz{\draw[->, thick] (0,0)--+(0.5,0);}
\end{minipage}
\begin{minipage}{0.20\textwidth}
\centering
\begin{forest}
for tree={s sep=10mm, inner sep=1, l=1}
 [$\sqcap$
    [$A$]
    [$\exists r$
      [$\sqcap$, , name=start, tikz={%
   \node [draw=red, thick, rounded corners, fit=() (endB) (endC), inner xsep=5pt, inner ysep=5pt] {};
   }
      [$\exists r$ [$C$, name=endC]]
      [$B$, name=endB]
    ]
  ]
  ]
\end{forest}
\end{minipage}}

\caption{Two possibilities for mutating the CE A $\sqcap\hspace{1pt}\exists$r. B.}
\label{fig:mutate-ce}
\end{figure}

\begin{figure}
\centering
\setlength{\tabcolsep}{17pt}
\scalebox{0.9}{
\begin{tabular}{cc}
\toprule
\textbf{Class Expression} & \textbf{Graphs} \\
\midrule
\begin{forest}
for tree={s sep=10mm, inner sep=1, l=1}
 [$\sqcap$
    [$A$]
    [$\exists r$
      [$\sqcap$
        [$B$]
        [$\exists r$
        [$\sqcap$
          [$B$]
          [$\exists r$
          [$A$]
          ]
          ]
        ]
      ]
    ]
  ]
\end{forest}&
\begin{tikzpicture}
  \node[circle, draw] (At) at (0.5,1) {A};
  \node[circle, draw] (Bt) at (2.5,1) {B};
  \node[circle, draw] (Ct) at (1.5,-0.732) {B};
  
  \draw (At) --node[above] {r} (Bt);
  \draw (Bt) --node[below right] {r} (Ct);
  \draw (Ct) --node[below left] {r} (At);
  \node[circle, draw] (A) at (0,2) {A};
  \node[circle, draw] (B) at (1,2) {B};
  \node[circle, draw] (C) at (2,2) {B};
  \node[circle, draw] (D) at (3,2) {A};
  
  \draw (A) --node[above] {r} (B);
  \draw (B) --node[above] {r} (C);
  \draw (C) --node[above] {r} (D);
\end{tikzpicture}
\\
\bottomrule
\end{tabular}}
\caption{Different graphs for the CE $\text{A} \sqcap\exists \text{r.} (\text{B} \sqcap \exists \text{r}. (\text{B} \sqcap \exists \text{r. A}))$.
}\label{fig:create-graph-from-ce}
\end{figure}
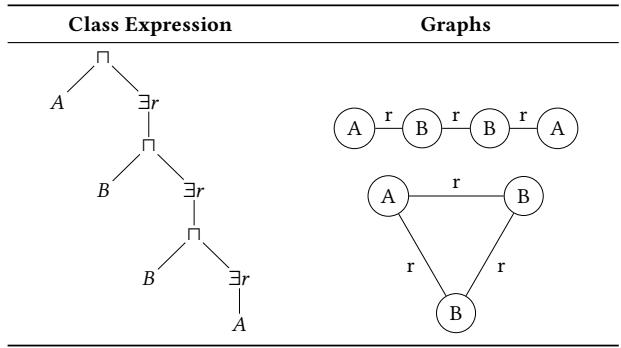

\paragraph{Mutate CE}
The mutation of CEs (line~8 of Algorithm~\ref{alg:beam-search}) is illustrated in Figure~\ref{fig:mutate-ce} and Algorithm~\ref{alg:mutate-ce}. The mutation function receives a CE $ce$ and a set of available $classes$. Like above, we create an expression $ce_{new}=\exists r.class_{new}$ where $class_{new}$ is chosen uniformly at random from $classes$ (lines~2--3 in Algorithm~\ref{alg:mutate-ce}). We add this expression $ce_{new}$ either to an existing intersection or an existing class, referred to as $ce_{old}$ (each with a probability of 50\%). In the former case, we uniformly randomly choose an intersection among all intersections $ce.intersections$ that appear in $ce$ and extend the intersection $ce_{old}$ with an additional operand (Figure~\ref{fig:mutate-ce} top, Algorithm~\ref{alg:mutate-ce} lines~6 and~10). In the latter case, we uniformly randomly choose a class $ce_{old}$ from the classes $ce.classes$ in the CE, and replace the class with a new intersection with the two operands $ce_{new}=\exists r. class_{new}$ and $ce_{old}$ (Figure~\ref{fig:mutate-ce} bottom, Algorithm~\ref{alg:mutate-ce} lines~8 and~10). 

\begin{algorithm}[tb]
\caption{Mutate a CE by adding a Class}\label{alg:mutate-ce}
\begin{algorithmic}[1]
\Procedure{MutateCE}{$ce, classes$}
\State $class_{new} \gets $ \textsc{Random}($classes$)
\State $ce_{new} \gets \exists r.class_{new}$
\State $place_{mutate} \gets$ \textsc{Random}(\{``intersection'', ``class''\})
\If{$place_{mutate} = $ ``intersection''}
    \State $ce_{old} \gets $ $ce$.\textsc{Random}($ce.intersections$)
\ElsIf{$place_{mutate} =$ ``class''}
\State $ce_{old} \gets $ $ce$.\textsc{Random}($ce.classes$)
\EndIf
\State $ce \gets$ Intersection($ce_{old}, ce_{new}$)
\State \textbf{return} $ce$
\EndProcedure
\end{algorithmic}
\end{algorithm}

\begin{algorithm}[tb]
\caption{Create Graph from CE}\label{alg:create-graph-from-ce}
\begin{algorithmic}[1]
    \Procedure{CreateGraph}{ce, graph=EmptyGraph()}
      \State $node \gets None$
    \If{\textsc{TypeOf}($ce$) is ``class''}
      \State $node \gets$ \textsc{CreateNodeOfType}($ce$)
      \State $graph$.\textsc{AddNode}($node$)
    \ElsIf{\textsc{TypeOf}($ce$) is ``intersection''}
      \State $class \gets \textsc{FindClass}(ce)$
      \State $node \gets$ \textsc{CreateNodeOfType}($class$)
      \State $graph$.\textsc{AddNode}($node$)
      \For{$op$ in $ce$ except for $op=class$}
        \State $ograph, onode \gets$ \textsc{CreateGraph}($op, graph$)
        \State $graph$.\textsc{MergeWithGraphOnNodes}(\par
        \quad\quad\quad ograph, onode, node)
    \EndFor
    \ElsIf{\textsc{TypeOf}($ce$) is ``existential restriction''}
      \State $fnode \gets None$
      \If{\textsc{RandomChoice()}}
       \State $fnode \gets graph$.\textsc{FindNode}($ce.filler$)
      \EndIf
      \If{$fnode$ is None}
        \State $fgraph, fnode \gets \textsc{CreateGraph}($\par
        \quad\quad\quad $ce.filler, graph)$
        \State $graph$.\textsc{AddGraph}($fgraph$)
      \EndIf
      \State $node \gets$ \textsc{CreateNodeWithoutType}()
      \State $graph$.\textsc{AddNode}($node$)
      \State $graph$.\textsc{AddEdge}($node$, $ce.property$, $fnode$)
    \EndIf
    \State \textbf{return} $graph$, $node$
    \EndProcedure
  \end{algorithmic}
\end{algorithm}

\begin{algorithm}[tb]
\caption{Scoring a CE}\label{alg:score-ce}
\begin{algorithmic}[1]
\Procedure{ScoreCE}{$ce, \lambda, GNN$, Aggr}
\State $graphs \gets EmptyList()$
\For{$1 \dots max\_number$}
\State $graph \gets$ CreateGraphs($ce$)
\State $v_r \gets$ RootNode($ce$)
\State $gnn\_outs$.append($GNN(graph, v_r)$)
\EndFor
\State $score \gets$ Aggr($gnn\_outs$) + $\lambda*$ Length($ce$)
\State $gnn\_out \gets GNN($
\EndProcedure
\end{algorithmic}
\end{algorithm}

\paragraph{Score CE} 
We propose maximizing the GNN output of graphs that are valid for a CE to score the CE. Towards this goal, we create graphs fulfilled by the CE and subsequently assess the GNN's performance on these graphs, as shown in Algorithm~\ref{alg:score-ce}. First, we explain the graph generation, then the scoring function.

We use Algorithm~\ref{alg:create-graph-from-ce} to create a graph from a CE recursively. This takes a CE and a graph as parameters. The graph is initialized as an empty graph when calling the function for the first time. In subsequent recursive calls, it contains the graph that has been constructed so far. The variable $node$ keeps track of the node that corresponds to the current CE's root class. In summary, Algorithm~\ref{alg:create-graph-from-ce} works as follows:

\begin{itemize}
\item If the CE consists solely of a named class (lines~3 to~5), then a graph with a single node of type $ce$ is created. 
\item If the CE is of type ``intersection'', e.g., the root element of the CE's parse tree is an intersection (lines~6 to~12), then the unique root class of $ce$ is determined through the function \textsc{FindClass} (line~7).%
\footnote{This is always possible due to how our CE creation and mutation works (see above). The construction ensures that each intersection has exactly one named class as an operand and an arbitrary number of existential restrictions. For example, our construction would create CEs like $A$, $A \sqcap \exists r. B$, $A \sqcap \exists r.B \sqcap \exists r.C$, but never $A \sqcap B$. The result of \textsc{FindClass($A \sqcap \exists r. B$}) would be $A$.}
Then a graph for each of its operands is created recursively (line~11) and is merged with the current graph (line~12) on their respective root nodes (root node $node$ of the current graph $graph$, and root node $onode$ of the operator graph $ograph$).
\item If the CE is an existential restriction (lines~13 to~22), we decide randomly with a probability of 50\% whether we attempt to reuse parts of the existing graph (line~16). If this is feasible, we identify a node for which the filler's CE is valid (line~17). Otherwise, we recursively create a new graph for the filler (line~20) and add it to the current graph (line~21). All that remains is to connect the filler graph to the current graph (lines 23--25). 
\end{itemize}

Figure~\ref{fig:create-graph-from-ce} gives an example by showing two possible graphs (right) generated from the CE (left). In the first example (top right), two different nodes of type A are created; in the second example (bottom right), a single node of type A is reused.
 
In the scoring function, we evaluate how well a given CE $ce$ explains the GNN's prediction of label $l$ by computing a score for the CE as follows:
\begin{enumerate}
        \item In each iteration, we generate a graph such that the CE $ce$ holds for the graph in the previously defined root node $v_r$, following the aforementioned procedure in Algorithm~\ref{alg:create-graph-from-ce} (line~4). Let $G(ce)$ be the set of generated graphs.
        \item For each graph $g \in G(C)$ we calculate the GNN prediction $f(g, v_g)$ (line~5) and aggregate these GNN predictions, using any aggregation function, like max or mean (lines~7 and~8).
\end{enumerate}
Thus, our provisional scoring function $\gamma$ of the CE $ce$ is
\begin{align*}
        \gamma(ce) = \aggr_{g\in G(ce)} f_l(g, v_g).
\end{align*}
The number of classes in one CE corresponds to its complexity. Hence, we propose regularizing the CE-scoring $\gamma$ by subtracting the number of classes in the CE, multiplied by the hyperparameter $\lambda$ (line~6), similarly to \citet{Heindorf2022EvoLearner}:
        \begin{align*}
                \text{score}(ce) = \gamma(ce) - \lambda \cdot \text{length}(ce).
        \end{align*}
In Section~\ref{sec:discussion}, we discuss further methods for regularization.

\subsection{Baseline}

As a baseline, we modify the approach by replacing the scoring function with a new objective: maximizing fidelity on a separate validation dataset. Consequently, we can omit the creation of graphs for each CE, resulting in explanations more closely tailored to the dataset on which the GNN is utilized.

\section{Evaluation}
We measured the quality of the generated explanations on a synthetic dataset using the metrics accuracy and fidelity.

\begin{table*}[t]
  \centering
  \setlength{\tabcolsep}{6pt}
   \caption{Three class expressions (CEs) as example results, sorted by their GNN score (top). For comparison, three CEs are retrieved by beam search maximizing fidelity on a separate validation dataset (middle), and three possible ground-truth CEs are given (bottom). For each CE, we evaluate fidelity, explanation accuracy (EA) and the maximal GNN output for graphs created by Algorithm~\ref{alg:create-graph-from-ce}. Fidelity was evaluated on a separate test dataset. We see, that each method reaches its goal optimizing GNN-output and fidelty, respectively.
  }
  \label{tab:BASpdfs}
  \textbf{Results on \textsc{Hetero-BA-Shapes}, for a GNN trained on 2 layers}
  \begin{tabular}{@{}l@{\hskip -115pt}cccr@{}}
    \toprule
    \textbf{CE scored by GNN output} 
    & \textbf{Class Expressions} 
    & \textbf{Fidelity} 
    & \textbf{EA}  
    & \textbf{GNN} 
    \\ 
    \midrule

$\text{1}^{\text{st}}$ 
 CE
&
    $
    B \sqcap \exists \text{to.} \left( \text{D}\sqcap  \exists\text{to.} \left( C \sqcap\exists \text{to.A} \sqcap \exists\text{to.B} \right)  \right)
    \sqcap \exists\text{to.C}
    \sqcap \exists\text{to.A}
    \sqcap \exists\text{to.D}
    \sqcap \exists\text{to.B}
    \sqcap \exists\text{to.A}
    \sqcap \exists\text{to.B}
    \sqcap \exists\text{to.D}  
    $
& $0.50$
& $0.45$  
& $12.52$ 
\\ 
  
    $\text{2}^{\text{nd}}$ 
     CE 
    &$
    B \sqcap \exists \text{to.} \left( \text{D} \sqcap \exists\text{to.B} \sqcap \exists\text{to.} \left( \text{C} \sqcap \exists\text{to.A}\right)    \right)
    \sqcap \exists\text{to.C}
    \sqcap \exists\text{to.A}
    \sqcap \exists\text{to.D}
    \sqcap \exists\text{to.B}
    \sqcap \exists\text{to.C}
    \sqcap \exists\text{to.B}
    $
    & $0.50$
    & $0.63$  
    & $12.39$
\\ 
  
    $\text{3}^{\text{rd}}$ 
    CE 
    &
    $
    B \sqcap \exists\text{to.} \left( \text{D} \sqcap \exists\text{to.B} \sqcap \exists\text{to.} \left( 
    C \sqcap \exists\text{to.A}       \right)      \right)
    \sqcap \exists\text{to.C}
    \sqcap \exists\text{to.A}
    \sqcap \exists\text{to.D}
    \sqcap \exists\text{to.B}
    \sqcap \exists\text{to.A}
    \sqcap \exists\text{to.B}
    \sqcap \exists\text{to.C}
    $
    & $0.51$ 
    & $0.63$  
    & $11.89$ 
    \\ \midrule
\textbf{CE scored by fidelity} &&&&
\\
\midrule
$\text{1}^{\text{st}}$ 
CE
&
    $
    \text{B} \sqcap \exists\text{to.} \left( \text{C} \sqcap \exists\text{to.} \left( \text{B} \sqcap \exists\text{to.B} \right) \sqcap \exists\text{to.C} \right)
    \sqcap \exists\text{to.A}
    $
& \textbf{0.97}
& $\textbf{1.00}$
&$0.40$ 
\\ 
  
    $\text{2}^{\text{nd}}$ 
    CE 
    &
    $
    \text{B} \sqcap \exists\text{to.} \left( \text{C} \sqcap \exists\text{to.} \left( \text{B} \sqcap \exists\text{to.B} \right) \sqcap \exists\text{to.C} \right)
    \sqcap \exists\text{to.A} \sqcap \exists\text{to.B} 
    $
    & $0.97$
    & $1.00$
    & $0.40$ 
\\ 
    $\text{3}^{\text{rd}}$ 
    CE 
    &
    $
    \text{B} \exists\text{to.} \sqcap \left(   
    \text{C} \sqcap \left( \text{C} \sqcap \exists\text{to.C} \sqcap \exists\text{to.} \left( \text{C} \sqcap \exists\text{to.B} \right) \right)
    \sqcap \exists\text{to.A}
    \sqcap \exists\text{to.B}
    \sqcap \exists\text{to.B}
    \sqcap \exists\text{to.C}
    \right)
    $
    & $0.97$ 
    & $1.00$  
    & $2.67$ 
    \\ \midrule
\textbf{Ground-truth CEs of increasing length} &&&&
\\
\midrule
$\text{1}^{\text{st}}$ CE
    &
    $B \sqcap 
       \exists \text{to.} \left(\text{B} \sqcap \exists \text{to.C}\right) \sqcap \exists \text{to.A} \sqcap \exists \text{to.C} 
      $
      & $0.93$ 
      & $1.00$ 
      & $-1.91$ 
\\ 
  
    $\text{2}^{\text{nd}}$ CE
    &
    $
    B\sqcap 
    \exists \text{to.} \left(\text{C} \sqcap \exists \text{to.C}\right) \sqcap
       \exists \text{to.} \left(\text{B} \sqcap \exists \text{to.C}\right) \sqcap \exists \text{to.} \left(\text{A} \sqcap \exists \text{to.B}\right) 
      $
    & $0.96$
    & $1.00$
    & $1.67$ 
\\ 
  
    $\text{3}^{\text{rd}}$ CE
    &
    $B\sqcap \exists \text{to.} \left(\text{A} \sqcap \exists \text{to.B} \right) 
    \sqcap
       \exists \text{to.} \left(\text{B} \sqcap \exists \text{to.A}
       \sqcap \exists \text{to.C}\right)
       \sqcap \exists \text{to.} \left(\text{C} \sqcap \exists \text{to.C} \right)
      $
    & $0.96$ 
    & $1.00$ 
    & $2.35$ 
    \\
    \bottomrule
  \end{tabular}
\begin{minipage}{0.4\textwidth}
  \begin{tabular}{@{}l@{\hskip -155pt}ccc@{}}
    \toprule
    \textbf{3 Graphs for: \hspace{3pt} CEs scored by GNN output} &  &  &   \\ \midrule
$\text{1}^{\text{st}}$ &
    \includegraphics[width=2cm]{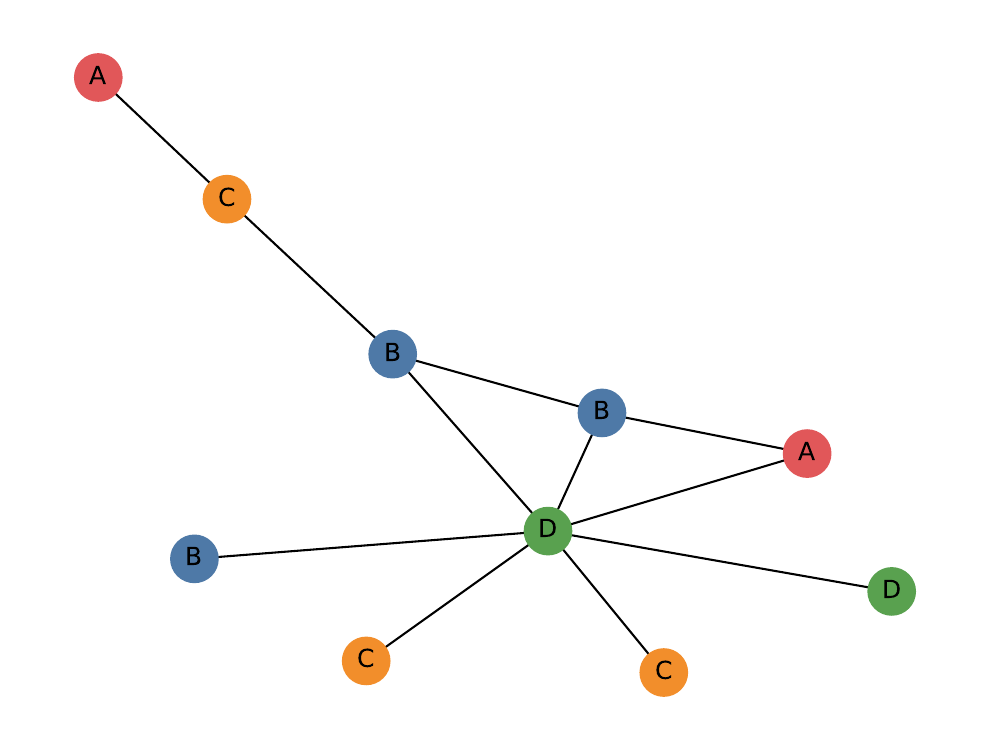} &
    \includegraphics[width=2cm]{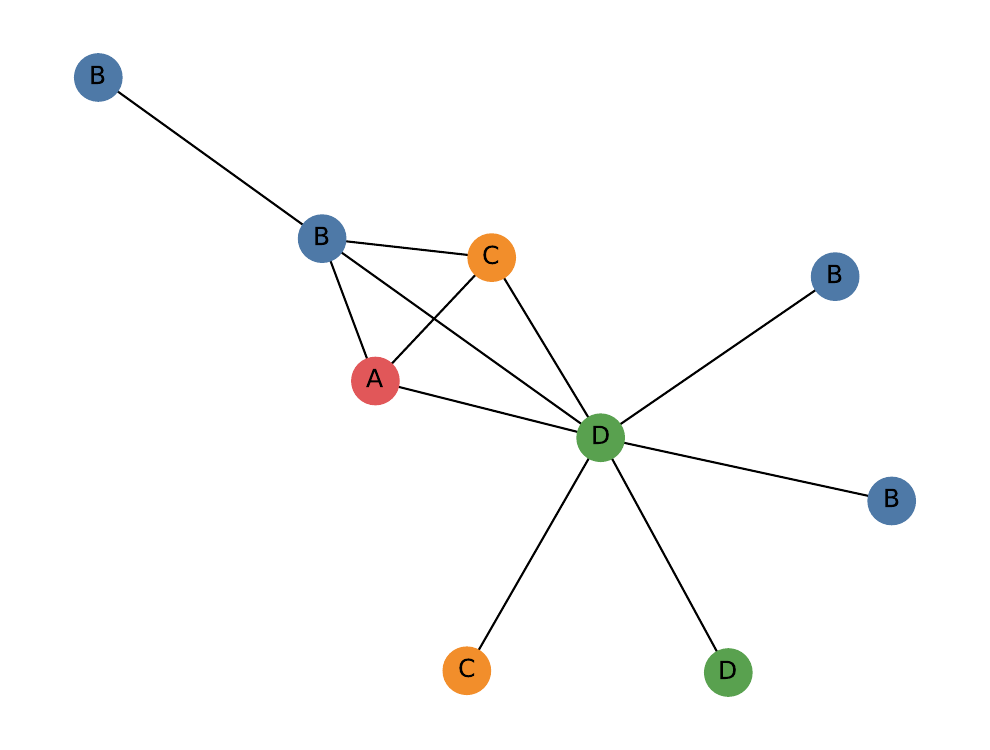} &
    \includegraphics[width=2cm]{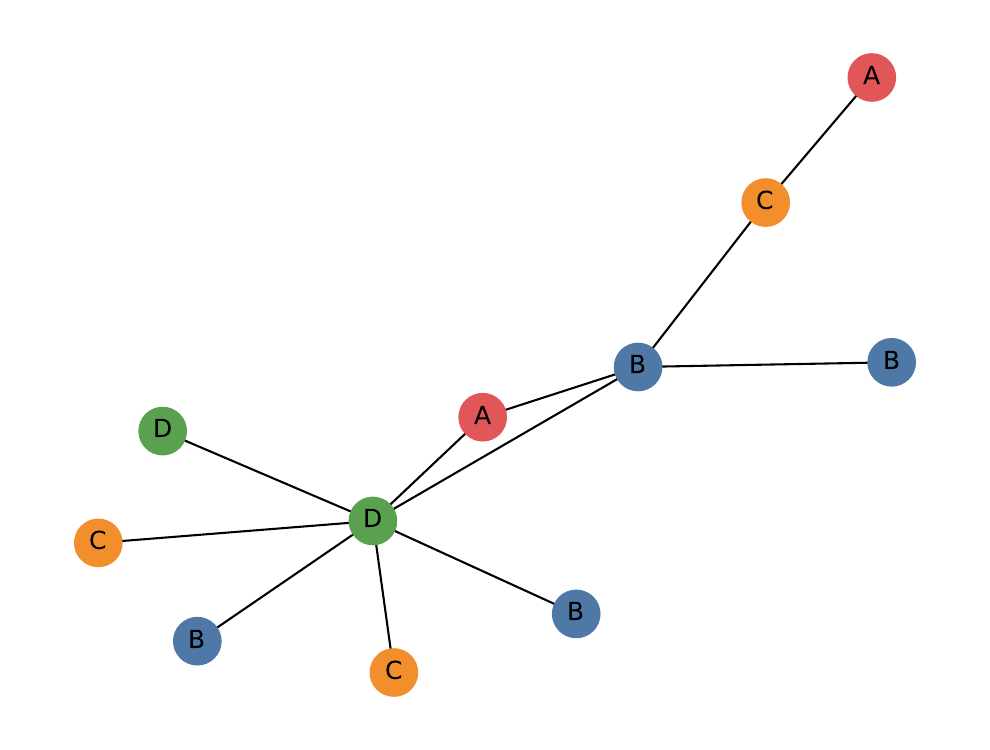} \\ 
    
    $\text{2}^{\text{nd}}$ &
    \includegraphics[width=2cm]{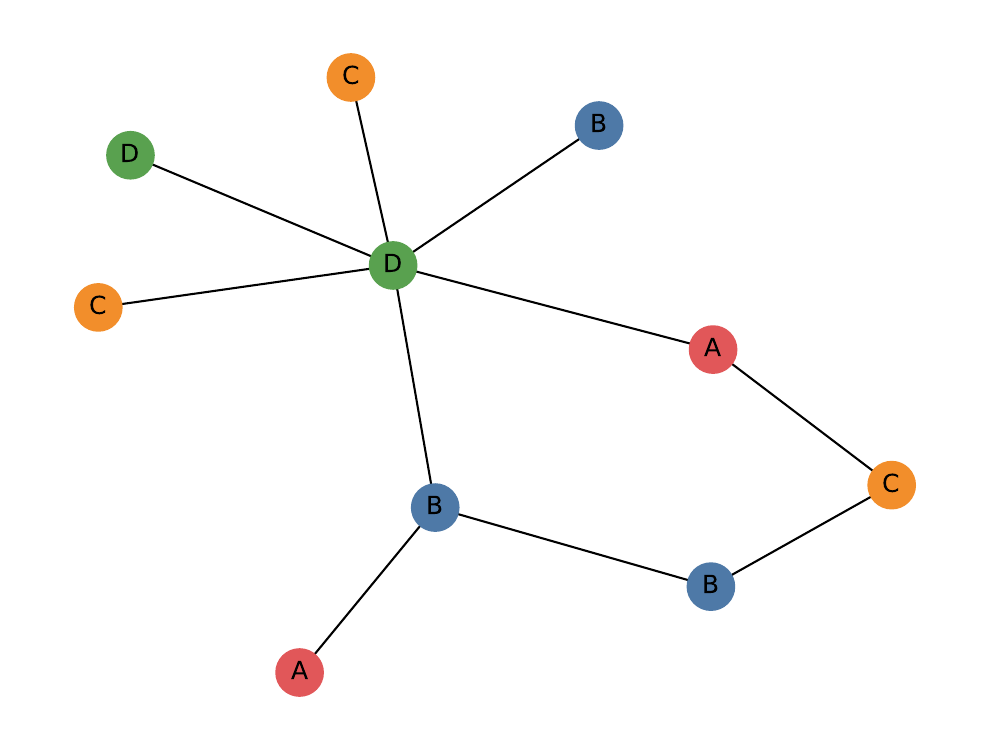} &
    \includegraphics[width=2cm]{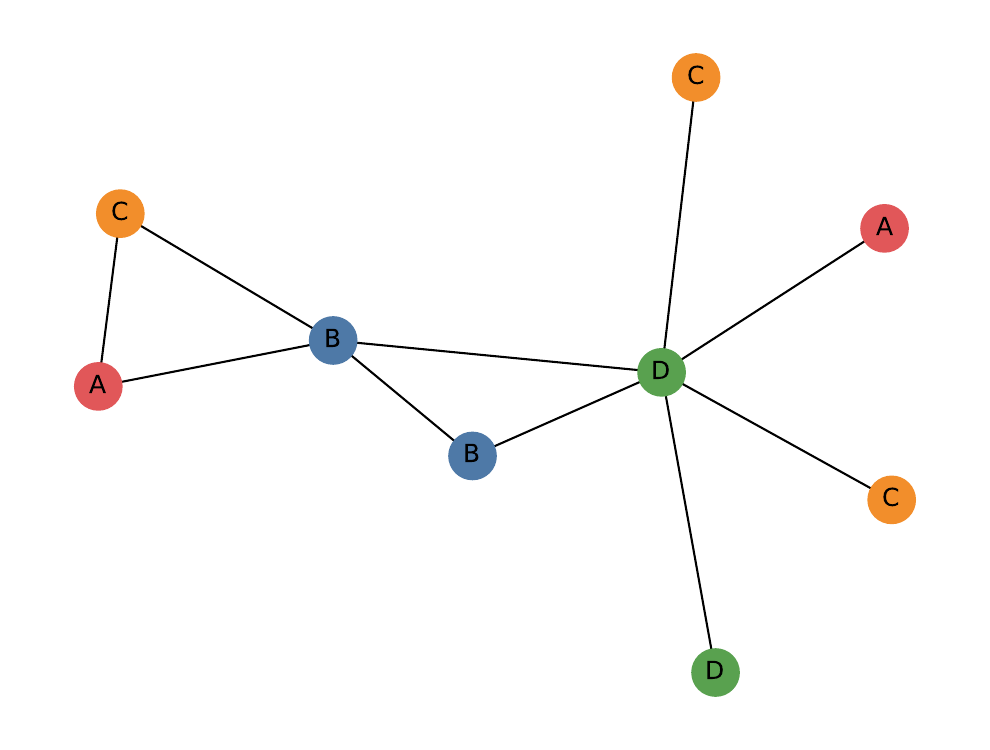} & 
    \includegraphics[width=2cm]{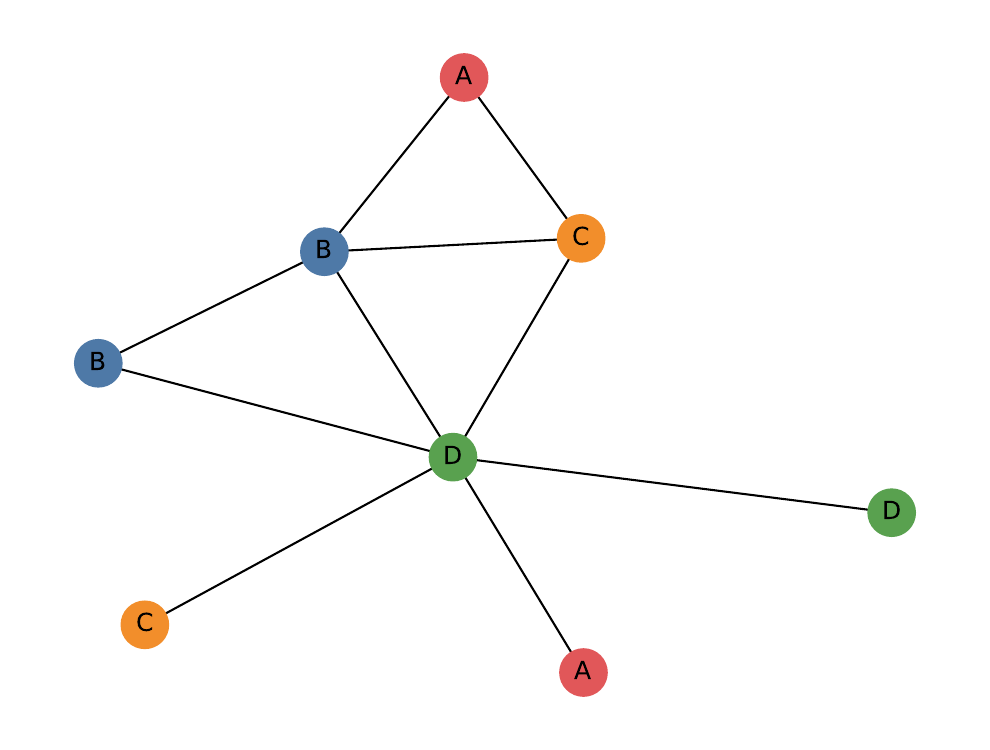} \\

    $\text{3}^{\text{rd}}$ &
    \includegraphics[width=2cm]{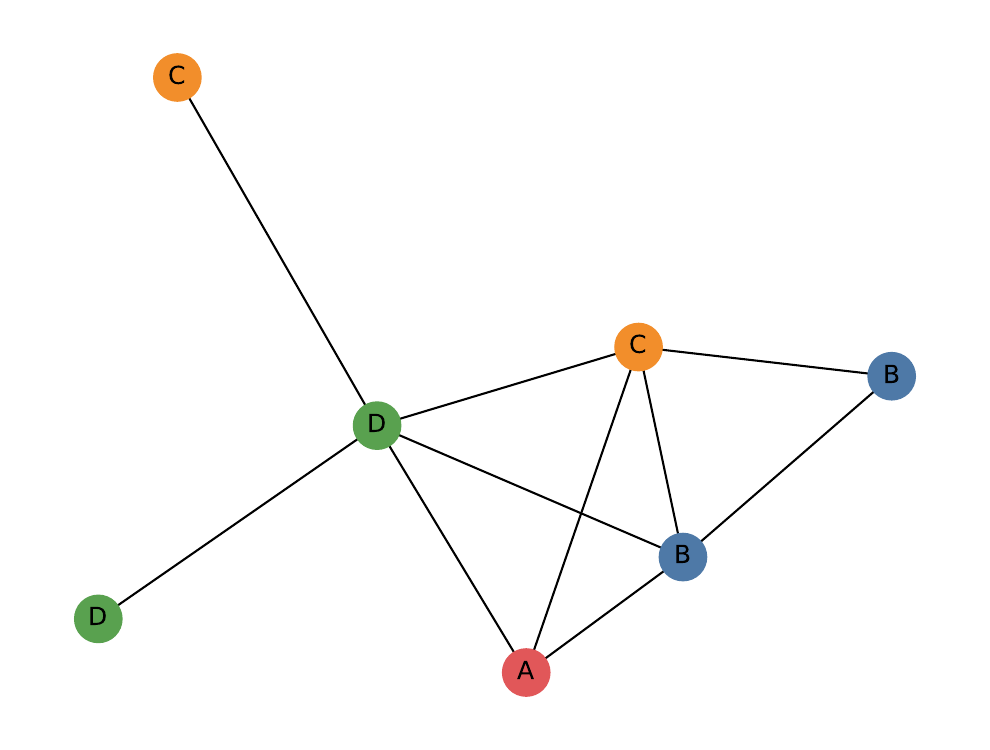} &
    \includegraphics[width=2cm]{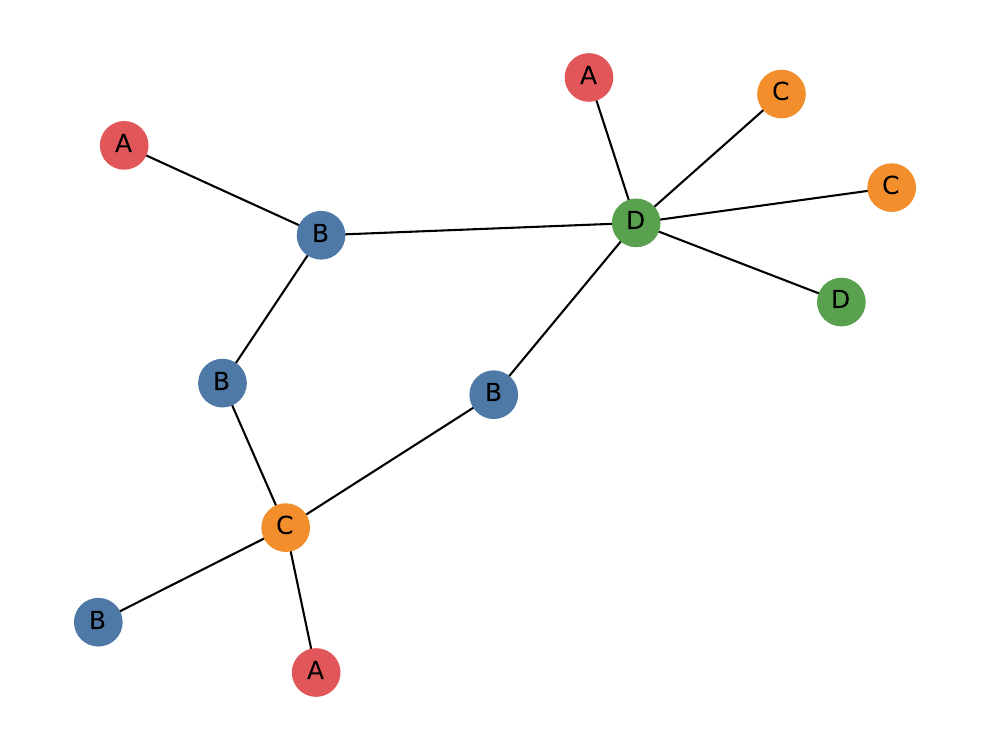} &
    \includegraphics[width=2cm]{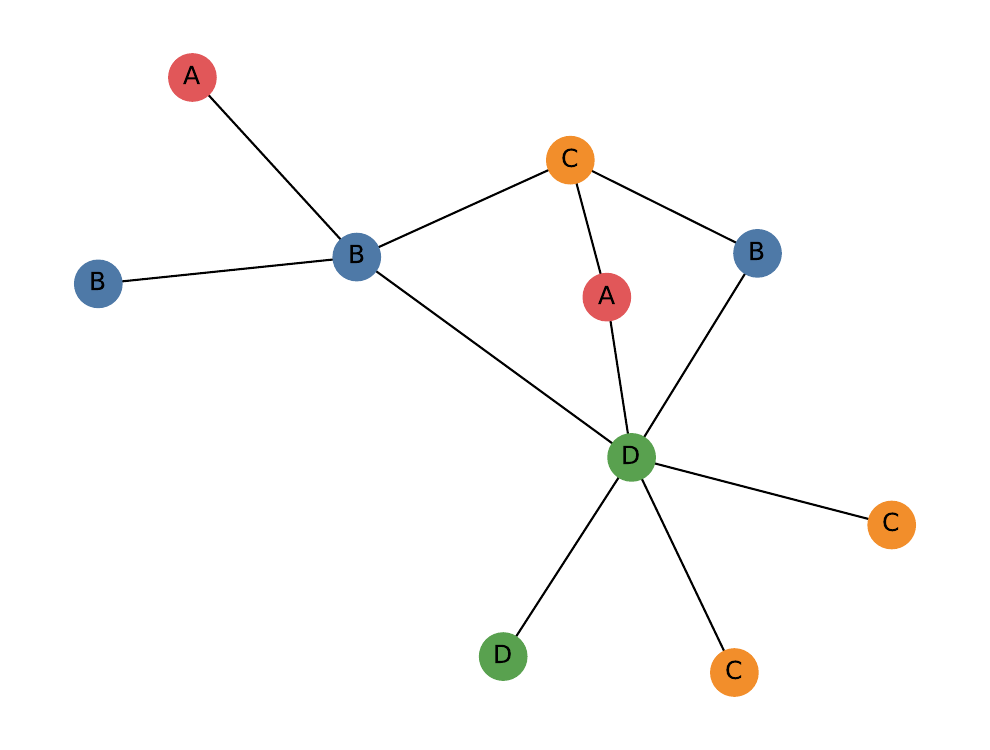}
    
   \\

    \bottomrule
  \end{tabular}
  \end{minipage}%
   \begin{minipage}{0.373\textwidth}
         \begin{tabular}{@{}l@{\hskip -20pt}cc@{}}
    \toprule
    \textbf{CEs scored by fidelity} & & \\ \midrule

    \includegraphics[width=2cm]{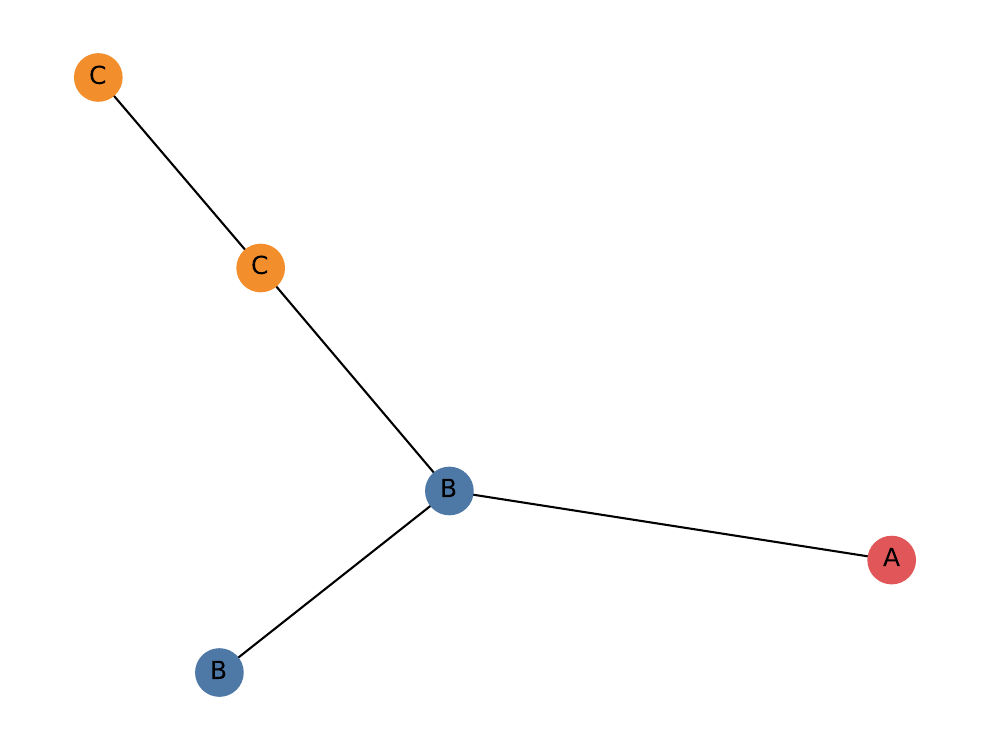} &
    \includegraphics[width=2cm]{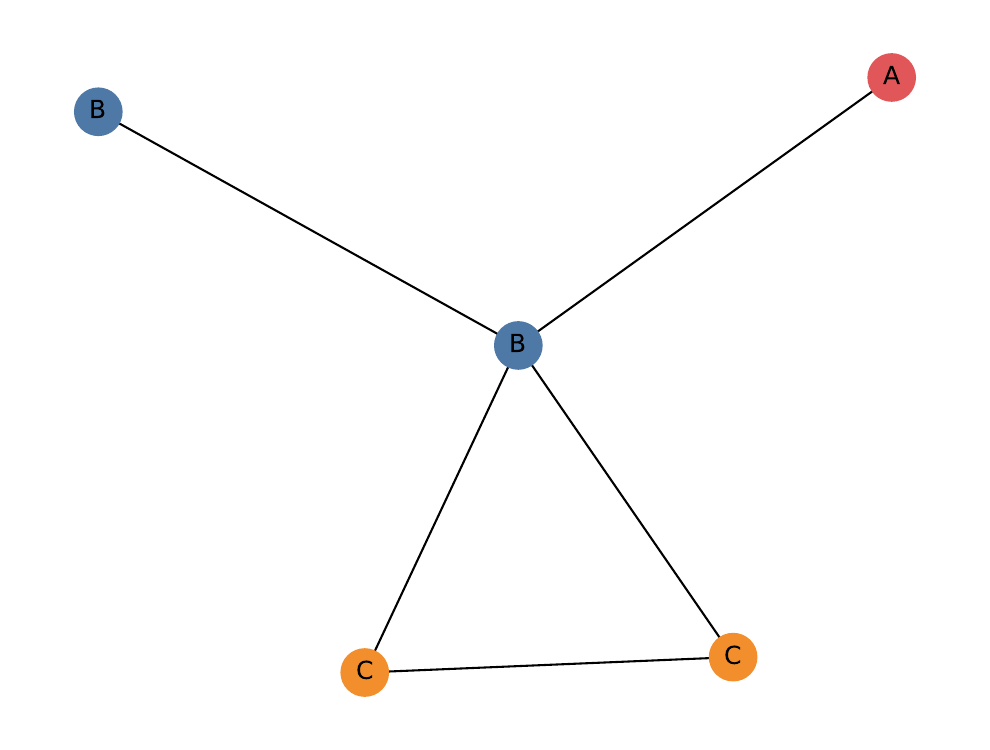} &
    \includegraphics[clip, trim=2cm 2cm 0cm 0cm, width=2cm]{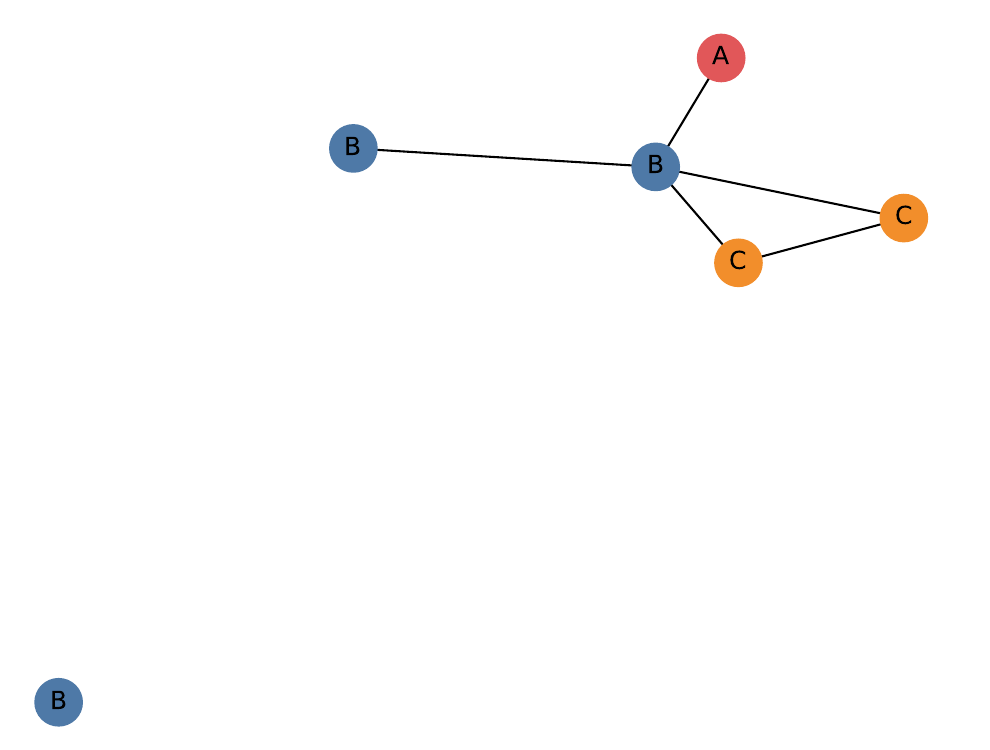}
    \\ 
    \includegraphics[width=2cm]{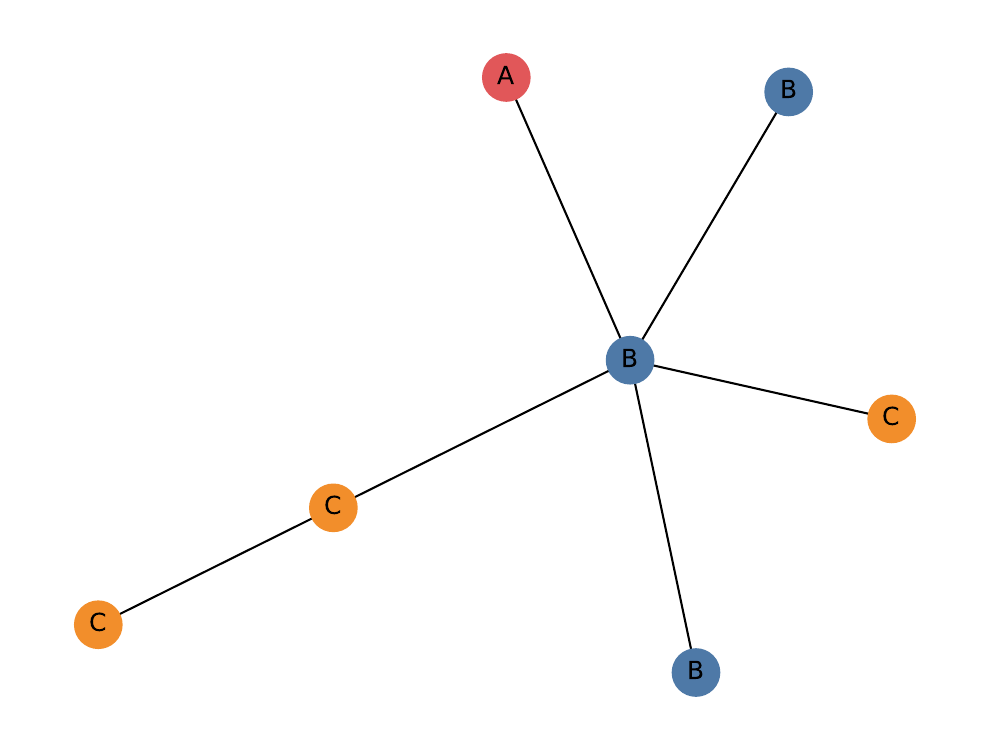} &
    \includegraphics[width=2cm]{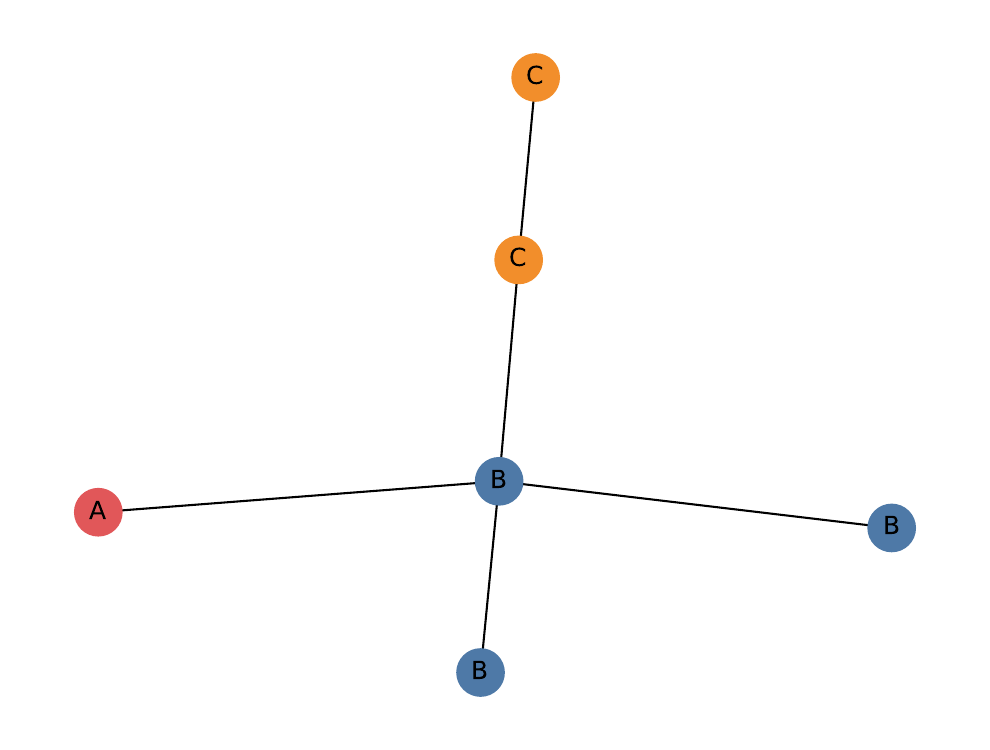} & 
    \includegraphics[width=2cm]{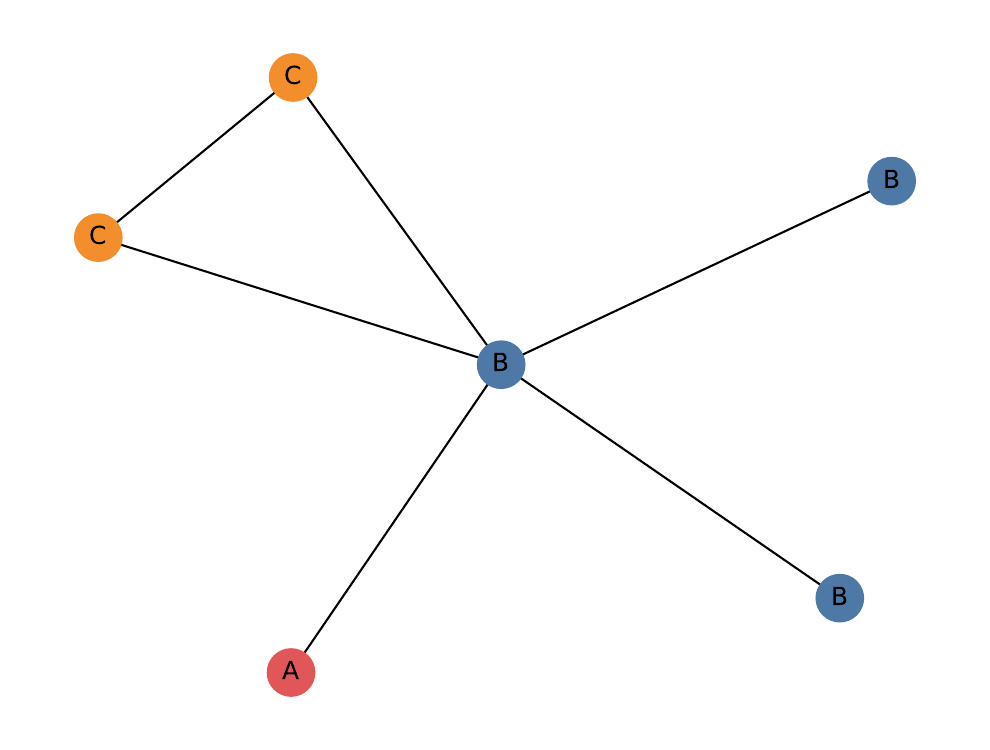}   
    \\
    \includegraphics[width=2cm]{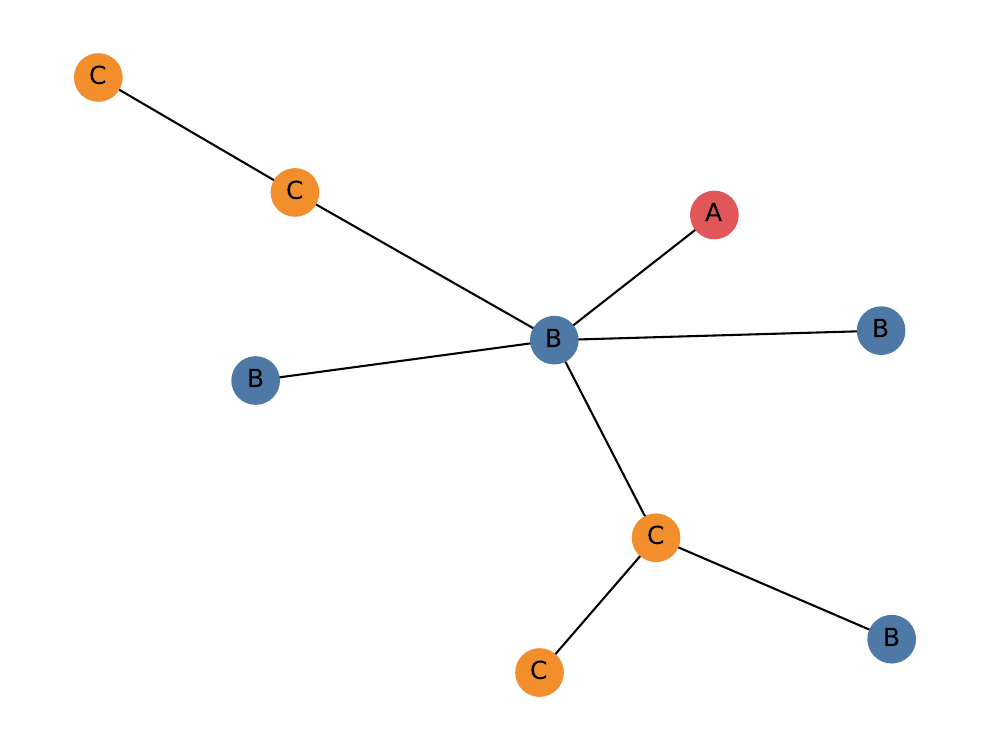} &
    \includegraphics[width=2cm]{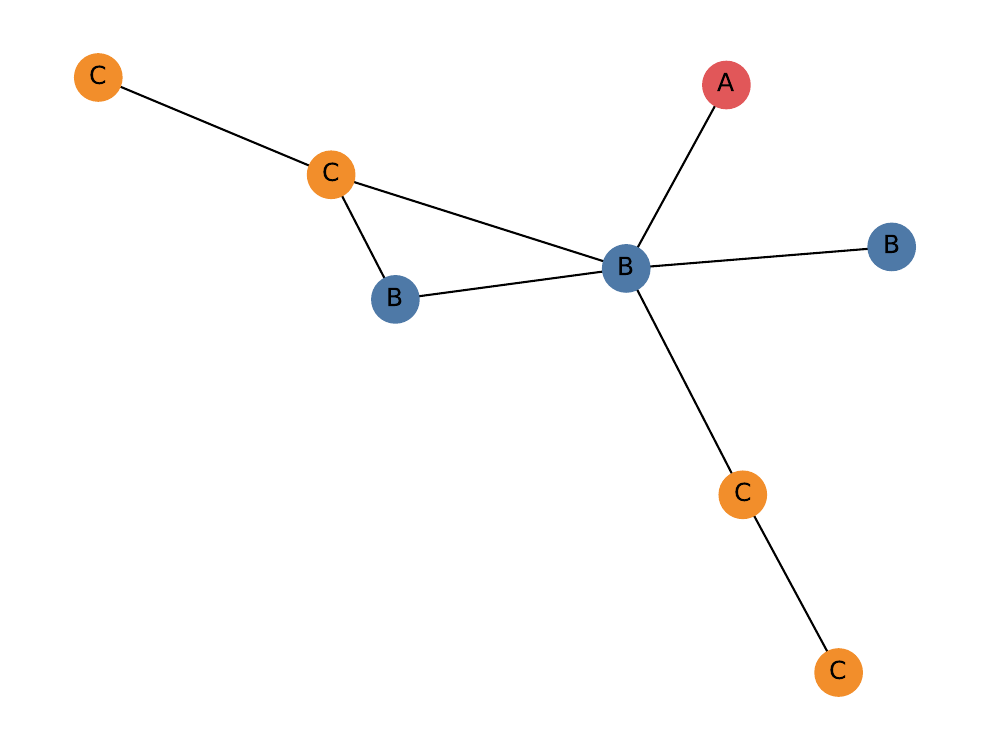} &
    \includegraphics[width=2cm]{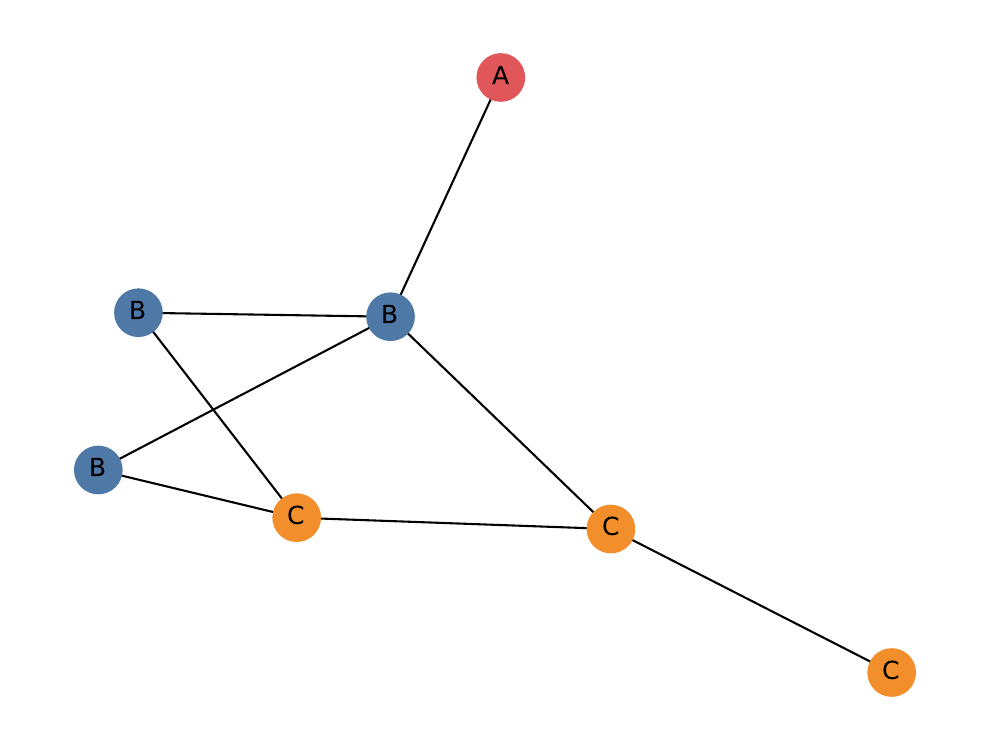} 
   \\
    \bottomrule
  \end{tabular}
   \end{minipage}%
   \begin{minipage}{0.35\textwidth}

   \begin{tabular}{@{}l@{\hskip +4.4pt}c@{}}
    \toprule
    \textbf{Graph for GT CEs} &\textbf{Legend} \\ \midrule

    \includegraphics[width=2cm]{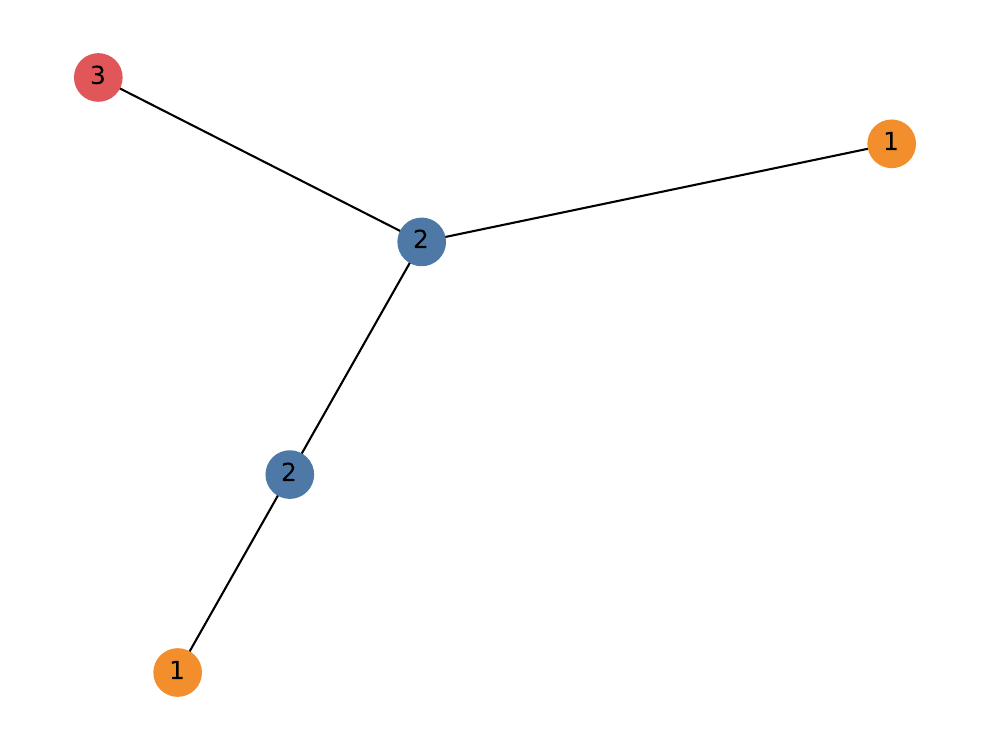} & \\ 

    \includegraphics[width=2cm]{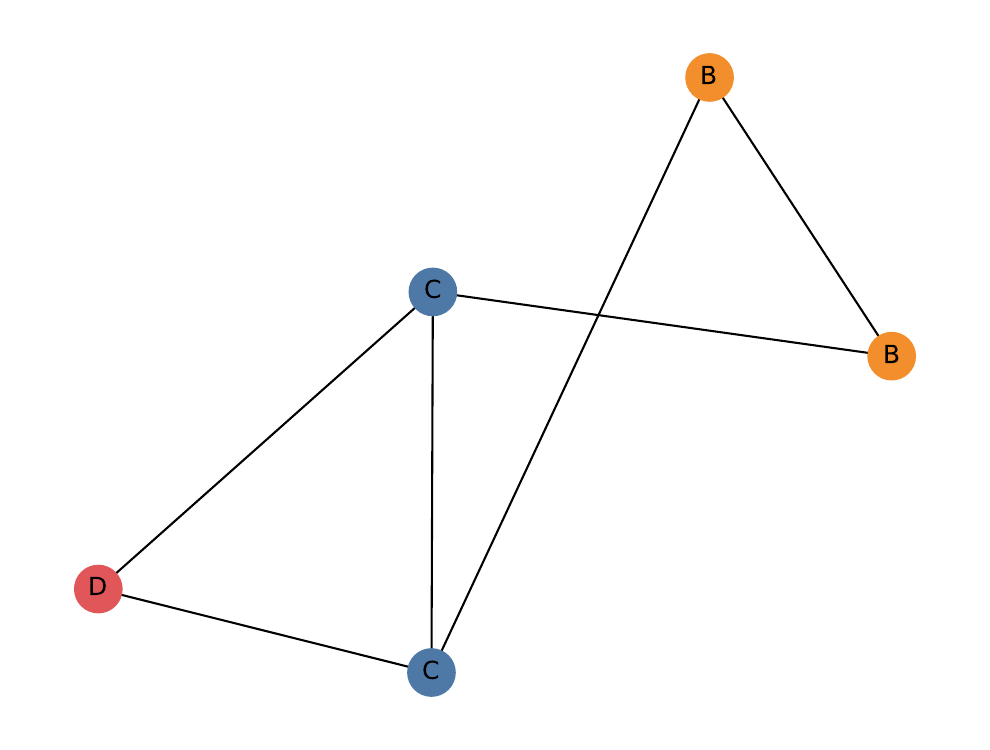} &
    \includegraphics[height=1.4cm]{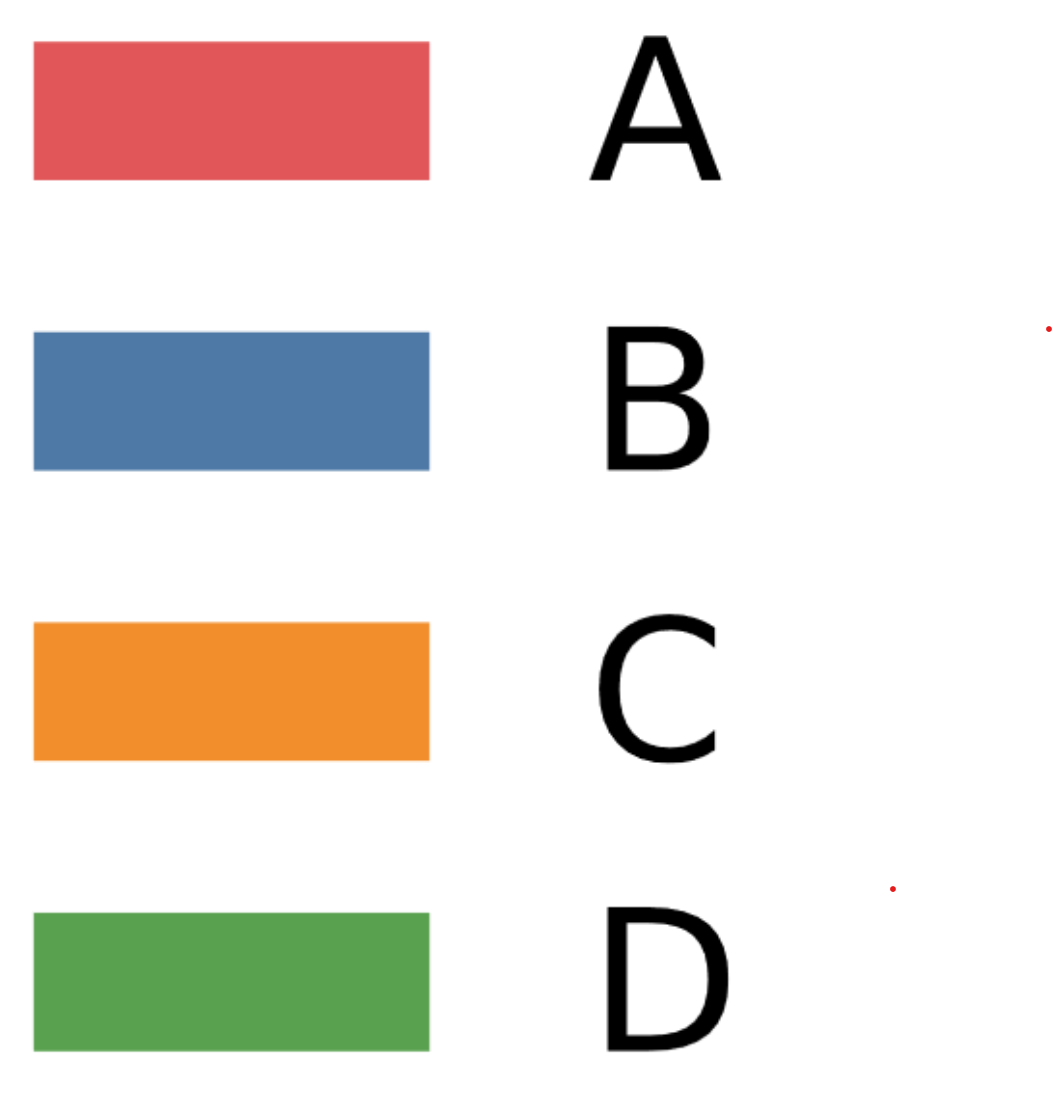}
    \\
    \includegraphics[width=2cm]{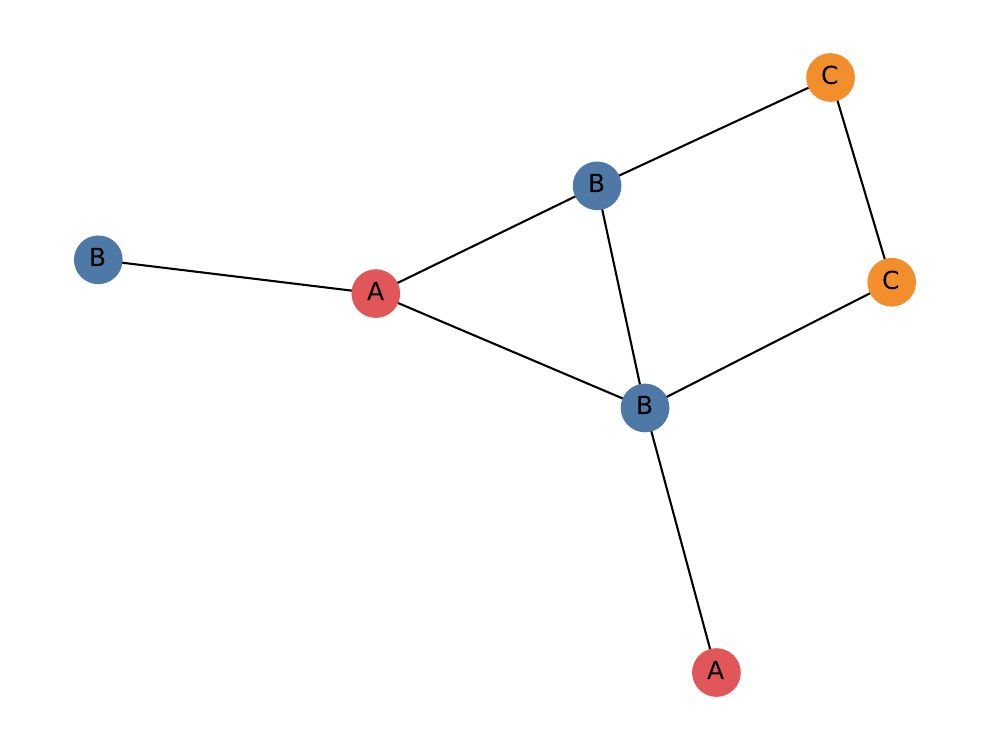} & \\
    \bottomrule
  \end{tabular}
  \end{minipage}
\end{table*}

\subsection{Evaluation Setup}

\paragraph{The \textsc{Hetero-BA-Shapes} Dataset}
We adjusted the \textsc{BA-Shapes} dataset~\cite{Ying2019GNNExplainer} to create a heterogeneous version. As in the \textsc{BA-Shapes} dataset, we first created a random Barabási-Albert Graph graph with $10\,000$ nodes, where each node is connected iteratively to the graph via $3$ edges; and then connected $1000$ house-motifs via a single edge to it. We introduced heterogeneity to the graph by assigning each edge to be of the named edge type \textit{to} and the nodes to different types based on their position: Each node at the top of the house motif was assigned to type $A$, the nodes in the middle of the house to type $B$, and the nodes at the bottom of the house to type $C$. Each node in the rest of the graph was assigned to type $A$, $B$, or $C$ by each $20\%$ probability, and of type $D$ by $40\%$ probability. Additionally, all nodes of type $B$ got the label $1$ if they are in a house motif, and $0$ otherwise. This results in $50\%$ of the nodes of type $B$ holding label $1$. At last, all nodes have one feature which was set to $1$.

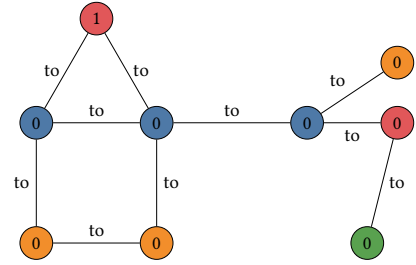
\begin{figure}
\centering
\scalebox{0.8}{
\begin{tikzpicture}
  \node[circle, draw, fill=HouseType1] (node1) at (0.5,-1) {0};
  \node[circle, draw, fill=HouseType1] (node2) at (2.5,-1) {0};
  \node[circle, draw, fill=HouseType2] (node3) at (0.5,1) {0};
  \node[circle, draw, fill=HouseType2] (node4) at (2.5,1) {0};
  \node[circle, draw, fill=HouseType3] (node5) at (1.5,2.732) {1};
  
  \node[circle, draw, fill=HouseType1] (node6) at (6.5, 2) {0};
  \node[circle, draw, fill=HouseType2] (node7) at (5,  1) {0};
  \node[circle, draw, fill=HouseType3] (node8) at (6.5,  1) {0};

  \node[circle, draw, fill=HouseType0] (node9) at (6,  -1) {0};
  
  \draw (node1) --node[above] {to} (node2);
  \draw (node1) --node[left] {to} (node3);
  \draw (node2) --node[right] {to} (node4);
  \draw (node3) --node[above] {to} (node4);
  \draw (node3) --node[left] {to} (node5);
  \draw (node4) --node[right] {to} (node5);

  \draw (node4) --node[above] {to} (node7);

  \draw (node6) --node[above left] {to} (node7);
  \draw (node7) --node[below] {to} (node8);

  \draw (node9) --node[right] {to} (node8);
  
\end{tikzpicture}}
\caption{One house motif (left) from our Hetero-BA-Shapes dataset that is connected to a random Barabasi-Albert Graph (right). The colors indicate the node type (red indicates type A, blue type B, orange type C, and green type D). The numbers in the nodes indicate the label. The red node on the right does not receive the label 1 because it is not part of a house motif.}
\label{fig:dataset}
\end{figure}

\paragraph{Training, Validation, and Test Sets} 
For training and validation of the GNN, we used one dataset with $15\,000$ nodes, consisting of $10\,000$ nodes not in a motif. For testing, we created a second dataset with $15\,000$ nodes. For creating explanations using the approach to maximize fidelity, we used one separate dataset of $15\,000$ nodes for evaluating fidelity. 

\paragraph{Ground-truth CEs} We compare the CEs retrieved from our approach and our baseline to different possible ground-truth CEs. We define a CE to be a \textit{ground-truth CE} to the house motif if a surjective homomorphism from the CE to the house motif exists. Three different ground-truth CEs are used, as shown in Table~\ref{tab:BASpdfs}: The first CE is defined as the shortest possible while still being a ground-truth CE.
The second CE contains six object properties corresponding to the six edges in the house motif. The third CE corresponds to the computation tree of a message passing GNN with two layers.

\paragraph{GNN Model}
We trained a 2-layer GNN on the Hetero-BA-Shapes dataset using one GraphSAGE layer~\cite{Hamilton2017GraphSAGE} per (node type, edge, node type)-triple and per layer. The GNN was trained for 1000 epochs with a training, validation, and test split of 40\%, 24\%, 36\%, respectively, using Adam-optimizer with a learning rate of $0.01$, reaching an accuracy of $0.97 \pm 0.1$ (single standard deviation) on the test set.

\paragraph{Hyperparameters for Explainer}
We performed beam search as in Algorithm~\ref{alg:beam-search}, using $k = 10\,000$ as beam width for $n = 10$ iterations. For each CE, we created $m=100$ graphs for scoring. For regularization, we used $\lambda = 0.5$. We chose the hyperparameter by iterative refinement, where minor modifications did not significantly alter the results.

\paragraph{Implementation} For our experiments, we used Python 3.8, with PyTorch 1.12, and PyTorch Geometric 2.01 for heterogeneous graphs, and the Ontolearn library for handling class expressions.%
\footnote{\href{https://pypi.org/project/ontolearn/}{https://pypi.org/project/ontolearn/}}

\subsection{Evaluation Metrics}
\label{sec:evaldef}

Our approach differs from most current approaches~\cite{Longa2022Explaining}, as we use CEs instead of subgraphs~\cite{Yuan2021SubgraphX} or motifs~\cite{Perotti2023Explaining} to explain the predictions. This necessitates slight modifications of existing evaluation metrics~\cite{Faber2021When,Amara2022GraphFramEx}, which we describe in the following.

\paragraph{Explanation Accuracy} We compute the explanation accuracy by mapping the CE to the ground-truth motif~\cite{Ying2019GNNExplainer,Faber2021When}, which, in the case of the \textsc{Hetero-BA-Shapes} dataset, is the house motif from Figure~\ref{fig:dataset}. For the computation, we extend the motif by one abstract node of type \textit{abstract}, which is connected to every node in the motif via every available edge type. We then map the CE to the motif by a homomorphism, starting with the CEs' root class. For node classification tasks, this root class is the node type we want to explain (in this paper $B$). Every class of the CE that cannot be mapped to a node in the motif is mapped to the abstract node. We count the true positives as all nodes in the motif a class was mapped on, the false negatives as the remaining nodes in the motif, and the false positives as the number of classes mapped on the abstract node. We don't use true negatives, as we only have positives in the motifs. If there are several possibilities for the mapping, we take the one leading to the highest accuracy: 
\begin{align*}
 \text{accuracy} = \frac{\text{tp}}{\text{tp}+\text{fp}+\text{fn}}.
\end{align*}
For example, the CE $\text{B} \sqcap \exists \text{to.A}$ has an accuracy of $0.4$, because it has two true positives (A and B) as both classes A and B from the CE are in the motif and each class can at most correspond to one tp, no false positives, and three false negatives ($C, C$, and $B$). As a more complex example, the accuracy of $\text{B} \sqcap \exists \text{to.}\left( \text{A} \sqcap \exists \text{to.C}\right)$ would be $\frac{2}{2+1+3}$, as there is no connection between nodes of type A and type C, leading to one false positive.

\paragraph{Fidelity}
Fidelity refers to the degree to which the explanation accurately reflects the model's behavior, i.e. whether it is possible to predict the GNNs' predictions from the explanation.
For calculating a CE's fidelity, we use CEs as binary classifiers, labeling every node that fulfills the CE with the label we want to explain and assigning all other nodes to another label (assuming binary classification). Then we use a test dataset, unseen by the GNN during training, to approximate fidelity as the similarity of the GNN and the CE, with $N$ being the number of test samples:
\begin{equation*}\label{eq:Fidelity3}
	\text{Fidelity} = \frac{1}{N} \sum_{i=1}^N \mathbbm{1}(y_{GNN}=y_{CE}).
\end{equation*}

\subsection{Experimental Results}
\label{sec:expres}

We observe the results in Table~\ref{tab:BASpdfs}, for CEs and their created graphs using Algorithm~\ref{alg:create-graph-from-ce}. We compare the results of our approach to the baseline approach, which maximizes fidelity, and three ground truth CEs of increasing length. We see, that the GNN scores highest on graphs, that don't contain the ground-truth house motif and even contain the (node type, edge type, node type)-triples, which aren't present in the house motif, as \textit{A, to, C} and \textit{D, to, D}. Removing these edges even decreases the GNN output on the graphs, leading to the conclusion that the GNN learned spurious correlations, as visualized in Table~\ref{tab:add_delete_edges}. The existence of spurious correlations on homogeneous motif datasets was additionally observed by~\cite{fan2023generalizing}. Furthermore, we notice that maximizing fidelity leads to CEs which can also serve as ground-truth CEs, hence correctly explaining the GNN. This suggests learning explanations by maximizing fidelity leads to better explanations, in contrast to maximizing the GNN score, as currently done in the literature for global explanations, as in~\cite{Yuan2020XGNN, Wang2023GNNInterpreter}. In the pipeline provided by~\cite{Azzolin2022Global}, both maximizing the GNN output and maximizing fidelity are used. In summary, the results show the ability to use our approach to detect spurious correlations and to explain GNNs with high fidelity.

\begin{table}[tbh]
\centering
\caption{The GNN scores with and without certain edge types are presented for the top-scoring graph created from the top-scoring CE in Table~\ref{tab:BASpdfs}. We expect the GNN score to decrease when edges from the motif are removed and to increase otherwise. However, we do not observe this increase in the GNN score, suggesting that the GNN has learned spurious correlations.} 
\label{tab:add_delete_edges}
\setlength{\tabcolsep}{10pt}
\begin{tabular}{@{}lcc@{}}
\toprule
\textbf{Edge-type} & \textbf{Edge in motif}
& \textbf{GNN Prediction} \\ \midrule
Original graph & & 12.52 \\ \midrule
w/o edges A--A & \text{\sffamily X} & $12.52$ \\
w/o edges A--B & \checkmark         &  $7.75$ \\
w/o edges A--C & \text{\sffamily X} & $11.67$ \\
w/o edges A--D & \text{\sffamily X} & $10.97$ \\
w/o edges B--B & \checkmark         &  $8.08$ \\
w/o edges B--C & \checkmark         &  $9.91$ \\
w/o edges B--D & \text{\sffamily X} &  $5.87$ \\
w/o edges C--C & \checkmark         & $12.52$ \\
w/o edges C--D & \text{\sffamily X} & $12.28$ \\
w/o edges D--D & \text{\sffamily X} & $12.12$ \\
\bottomrule

\end{tabular}
\end{table}

\section{Discussion}
\label{sec:discussion}

\paragraph{Detecting spurious correlation}
With our method, we can detect (node type, edge type, node type) triples the GNN focuses on, making it easier to debug GNNs.

\paragraph{Heterogeneous vs. homogeneous datasets} 
By delivering more expressive explanations, especially for datasets with different node and edge types, we encourage the use of heterogeneous GNNs instead of homogeneous ones, resulting in more expressive explanations while still achieving similar results~\cite{wang2023enabling}. The approach can also be applied to homogeneous datasets where categorical node features can serve as classes instead of node types.

\paragraph{Maximizing GNN-score vs. maximizing fidelity}
Our results indicate that learning explanations using metrics as fidelity creates explanations for the general behavior of the model while maximizing the GNN output identifies outliers. This is because fidelity considers every node classified under a desired label by the GNN, providing a comprehensive view, whereas maximizing GNN output focuses only on a potentially non-representative subset.

\paragraph{Model-specific vs. model-agnostic}
Our approach is mod\-el-ag\-nos\-tic, as it treats the GNN as a black box. This has the advantage that our approach can generate explanations for all architectures of GNNs.

\paragraph{Node classification vs. graph classification} Although we focus on node classification, the approach is extendable to graph classification. In this case, we learn a CE by scoring the GNN on the created graphs. If any node in the graph fulfills the CE, the entire graph is classified accordingly, yielding the corresponding label.

\paragraph{Runtime}
The algorithm runtime is primarily determined by the graph-creation step during scoring, which takes significantly longer for longer CEs. For GNNs with more layers, the explanations become longer, and larger graphs must be created for scoring as additional layers make the GNN take a larger neighborhood into account. Therefore, the algorithm scales to the complexity of the GNN.

\begin{figure}
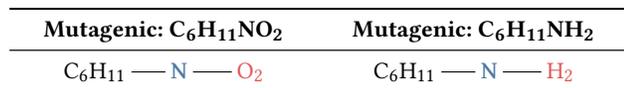

\setlength{\tabcolsep}{13pt}
\begin{tabular}{cc}
\toprule
\textbf{Mutagenic: \ce{C6H11NO2}} & \textbf{Mutagenic: \ce{C6H11NH2}} \\
\midrule
\chemfig{C_6H_{11}-{\color{HouseType2}\mathbf{N}}-{\color{HouseType3}O_2}}
&
\chemfig{C_6H_{11}-{\color{HouseType2}\mathbf{N}}-{\color{HouseType3}H_2}}
\\
\bottomrule
\end{tabular}
\caption{Two nitrogen nodes (left and right in blue) predicted to be part of a mutagenic molecule. Previous explanation approaches would either identify the graph pattern \ce{N}--\ce{O_2} or \ce{N}--\ce{H_2} as mutagenic, but they were unable to express both explanations at the same time. In contrast, the class expression $\ce{N}\sqcap \exists \text{hasBond. } (\ce{H2} \sqcup \ce{O2})$ can do so. Example adopted from~\citet{Luo2020PGExplainer}.
}
\label{fig:MUTAG}
\end{figure}

\paragraph{Applications}
This approach may be utilized for explaining molecule datasets, as promoted in Figure~\ref{fig:MUTAG}, helping researchers gain new insights into the data by revealing previously undetected structures. Furthermore, our method can be used for debugging a trained GNN, by finding GNN peaks outside the training data and detecting spurious correlations as discussed in Section~\ref{sec:expres}, leading to more explainable and transparent GNNs.

\paragraph{Regularization} 
The maximal node degree can be regularized to create less dense explanations or explanations more aligned with the training data. Additionally, regulating the graph distance avoids long chains of explanations.

\paragraph{Features} We propose introducing features as data properties into our explanation by leveraging techniques from explainable AI on tabular data, especially decision trees.

\paragraph{Comparison with other explanation approaches} In future work, it might be interesting to compare our explanations to further approaches creating rule-based explanations, like EvoLearner~\cite{Heindorf2022EvoLearner}, GLGExplainer~\cite{Azzolin2022Global} or the approach of Himmelhuber et al.~\cite{Himmelhuber2021Combining}. Additionally, a comparison to graph-based explanations like the motifs generated by Perotti et al.~\cite{Perotti2023Explaining} or the heterogeneous implementations of Subgraphx~\cite{Yuan2021SubgraphX}\footnote{\url{https://docs.dgl.ai/en/1.1.x/generated/dgl.nn.pytorch.explain.HeteroSubgraphX.html}}, PGExplainer~\cite{Luo2020PGExplainer} \footnote{\url{https://docs.dgl.ai/en/latest/generated/dgl.nn.pytorch.explain.HeteroPGExplainer.html}} and GNNExplainer~\cite{Ying2019GNNExplainer}\footnote{\url{https://docs.dgl.ai/en/1.1.x/generated/dgl.nn.pytorch.explain.HeteroGNNExplainer.html}} could lead to interesting discoveries.

\paragraph{ALCQ} In our forthcoming research, we aim to refine our methodology by incorporating the description logic $\mathcal{ALCQ}$, with a specific focus on unions and cardinality constraints. Figure~\ref{fig:MUTAG} illustrates how unions can enhance the expressivity of explanations. A challenge associated with using unions is ensuring that no information is duplicated across two separate operators within a single union. To adapt our graph creation process, we plan to limit edge connections in the graph such that they only connect with newly created nodes. This adjustment aligns with the GNN's computation graph and prevents ambiguity, consistent with the message-passing procedure employed by most GNNs.

\section{Conclusion}
This paper presents a novel approach for explaining GNNs globally through class expressions from description logics. By this, we identify and tackle two problems of XAI on graph-structured data, (1)~current graph-based explanations do not provide global explanations and 
(2)~maximizing the GNN is less suitable for finding global explanations than evaluating the explanations on a validation dataset. We demonstrated the effectiveness of our approach on heterogeneous graphs on a synthetic dataset. In Section~\ref{sec:discussion}, we give an extensive outlook on possibilities for future work. Overall, this paper provides a foundation for utilizing class expressions as explanations for heterogeneous GNNs, contributing to the field of XAI on graph-structured data.

{
\bibliographystyle{ACM-Reference-Format}
\raggedright
\bibliography{arxiv24-xgnn-logics-lit}
}

\end{document}